\documentclass{article} % For LaTeX2e
\usepackage{iclr2025_conference,times}

\usepackage{amsmath,amsfonts,bm}

\def\eqref#1{equation~\ref{#1}}
\def\1{\bm{1}}

\DeclareMathAlphabet{\mathsfit}{\encodingdefault}{\sfdefault}{m}{sl}
\SetMathAlphabet{\mathsfit}{bold}{\encodingdefault}{\sfdefault}{bx}{n}

\usepackage{fancyhdr}

\usepackage{hyperref}
\usepackage{url}
\usepackage{nicefrac}
\usepackage{float}
\usepackage{placeins}
\restylefloat{table}
\usepackage{arabtex}
\usepackage{longtable}
\usepackage{utf8}
\setcode{utf8}
\usepackage{colortbl}
\usepackage{graphicx}
\usepackage[a4paper,margin=1in]{geometry}
\usepackage{multirow}
\usepackage{comment}
\usepackage{svg}
\usepackage{booktabs, caption, subcaption}
\usepackage{framed}
\usepackage{wrapfig}
\usepackage[most]{tcolorbox}
\usepackage{mdframed}
\usepackage[normalem]{ulem}
\usepackage{multirow}
\usepackage{makecell,array,adjustbox}

\title{Tahakom LLM Guidelines and Recipes: From Pre-Training Data to an Arabic LLM}

\author{
    \textbf{Areej AlOtaibi}\thanks{Equal contribution} \quad
    \textbf{Lina Alyahya}\footnotemark[1] \quad
    \textbf{Raghad Alshabanah}\footnotemark[1] \quad
    \textbf{Shahad Alfawzan}\footnotemark[1] \quad
    \textbf{Shuruq Alarefei}\footnotemark[1] \quad\\[0.5em]
    \textbf{Reem Alsabti}\footnotemark[1] \quad
    \textbf{Nouf Alsubaie}\footnotemark[1] \quad
    \textbf{Abdulaziz Alhuzaymi}\footnotemark[1] \quad
    \textbf{Lujain Alkhelb}\footnotemark[1] \quad
    \textbf{Majd Alsayari}\footnotemark[1]\quad
    \textbf{Waad Alahmed}\footnotemark[1]\\[0.8em]
    \textbf{Omar Talabay} \quad
    \textbf{Jalal Alowibdi} \quad
    \textbf{Salem Alelyani} \quad
    \textbf{Adel Bibi}\thanks{University of Oxford}
}

\iclrfinalcopy % Uncomment for camera-ready version, but NOT for submission.

\begin{document}

\maketitle

\begin{figure}[h]
    \centering
    \includegraphics[width=0.12\linewidth]{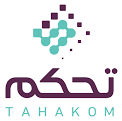}
    \label{fig:your-label}
\end{figure}

\begin{abstract}
Large Language Models (LLMs) have significantly advanced the field of natural language processing, enhancing capabilities in both language understanding and generation across diverse domains. However, developing LLMs for Arabic presents unique challenges. This paper explores these challenges by focusing on critical aspects such as data curation, tokenizer design, and evaluation. We detail our approach to the collection and filtration of Arabic pre-training datasets, assess the impact of various tokenizer designs on model performance, and examine the limitations of existing Arabic evaluation frameworks, for which we propose a systematic corrective methodology. To promote transparency and facilitate collaborative development, we share our data and methodologies, contributing to the advancement of language modeling, particularly for the Arabic language. 
\end{abstract}
\section{Introduction}

Large Language Models have evolved into powerful and versatile tools, revolutionizing a broad spectrum of fields, from the technical foundations of AI and computer science to practical applications in healthcare \cite{healthcare}, finance\cite{finance}, education\cite{education}, and beyond \cite{llms_survey}. By leveraging vast datasets that encompass diverse domains such as text, code, and mathematical equations, LLMs demonstrate exceptional abilities in comprehending, generating, and transforming human language.

Despite these advancements, the development of powerful open-source LLMs has largely centered around the English language\cite{open_llms}, limiting their relevance in other linguistic and cultural contexts. In the Arabic language domain, encouraging progress has been made with the release of models such as Allam\cite{allam_paper}, Fanar\cite{team20fanar25fanar}, Aya\cite{aya}, AceGPT\cite{acegpt}, and Jais\cite{jais}.  These models have shared their weights and training recipes, helping to expand the Arabic LLM ecosystem. However, to enable true reproducibility and sustained research progress, broader transparency remains essential, particularly through the release of code repositories, detailed documentation of datasets and their sources, as well as access to in-house evaluation benchmarks and training data. For instance, Jais has provided a comprehensive overview of its data sources, while Aya has gone further by making its entire dataset publicly available, offering a valuable resource for the research community. Despite these contributions, the overall availability of Arabic LLM resources remains sparse compared to their English counterparts.

The limited number of Arabic LLMs, combined with the often incomplete nature of their open-sourcing, poses a significant challenge for the field. Most models lack access to full training pipelines, datasets, or detailed documentation of optimization techniques, making it difficult for researchers to replicate results, analyze model behavior, or extend existing work. Advancing Arabic LLM research requires a stronger commitment to openness and consistency, inspired by the openness that has accelerated innovation in English language LLMs.

In this work, we outline the research and development efforts focused on building an Arabic LLM, emphasizing the various stages of its development. This includes data curation, pre-training data refinement through filtration experiments, tokenization strategies, evaluation and benchmarking. We explore how these choices differ from the development of English LLMs and discuss the challenges faced during the process, as well as the improvements implemented. The aim is to provide valuable insights into advancing the capabilities of Arabic LLMs. Our contributions can be summarized in three folds:

\begin{itemize}
    \item We construct and release a high-quality Arabic pre-training dataset from Common Crawl using a multi-stage pipeline involving extraction, language identification, heuristic and model-based filtering, and de-duplication. The pipeline is validated through ablation experiments, and the final dataset will be publicly released to support future research on Arabic LLMs. 
    %\bibi{need to be fixed}
    \item We conduct an empirical study to evaluate and quantify the impact of tokenizer training choices such as Vocabulary size, training data composition, and Pre-Tokenization methods on the downstream performance of LLMs.
    \item We improve the evaluation of Arabic language models by a refined and modified benchmark like ARB-MMLU delivers more dependable assessment than current translated datasets, as introducing culturally relevant evaluation data and establishing a comprehensive framework for systematic model assessment.

\end{itemize}

% \newpage

\begin{figure}[t]
    \centering
    \includegraphics[width=1\textwidth]{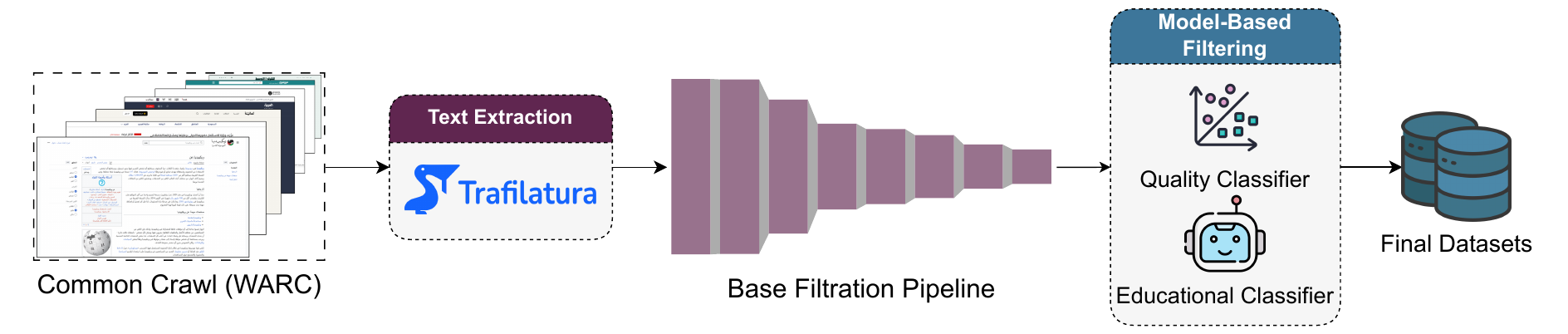}
    % \vspace{-0.15cm}
    \caption{Overview of the data preparation pipeline, from data extraction to model-based filtering. WET files provide ready-to-use plain text, whereas WARC files require text extraction. After that, we apply the base filtration pipeline,  followed by model-based filtering to obtain high-quality Arabic Pre-training data.}
    \label{fig:overall pipeline}
\end{figure}

\section{Pre-Training}
\label{sec:Pre-training}
\renewcommand{\arraystretch}{1.5} % 

Pre-training is the foundational stage in developing a large language model. During this stage, the model is exposed to vast amounts of language data, allowing it to learn the structure, semantics, and patterns of the language. LLMs typically require massive volumes of high-quality data, which can be difficult to obtain. The challenge becomes even greater when targeting languages with a limited online presence, such as Arabic: it is the sixth most-spoken language worldwide (Central Intelligence Agency 2025), yet appears in only about 0.6\% of pages in the first two Common Crawl releases of 2025 (Common Crawl 2025). Public corpora such as 101 B Arabic Words \cite{aloui2024101}, ArabicWeb24 \cite{ArabicWeb24}, and the Arabic slice of FineWeb2 \cite{penedo2025fineweb2} mitigate the shortage to some extent, yet their scale remains modest, leaving a persistent gap for large-scale, high-quality Arabic training data. Closing this gap calls for broader, better-documented Arabic datasets and foundation models that can support sustained research progress.

\subsection{Pre-Training Arabic Data: Common Crawl}

Common Crawl \cite{commoncrawl} is the largest open-source web crawling project and stands as one of the most critical data sources for training LLMs. It serves as the foundational source for several widely-used datasets, such as FineWeb \cite{penedo2024finewebdatasetsdecantingweb}, DataComp \cite{li2024datacomplmsearchgenerationtraining}, and RedPajama-Data-v2 \cite{weber2024redpajamaopendatasettraining}, known for their high-quality and comprehensive scale in data collection and filtration within the open-source community.
Common Crawl provides two formats: 1- \textbf{WARC} files, which store raw content from web pages, including HTML tags, JavaScript code, and extensive metadata. This format is great when tailored content extraction methods are required. However, WARC files are computationally heavy; since they contain raw content, the file sizes are large, and combined with the need for extraction, this highlights the scaling challenges.
2- \textbf{WET} files, which contain plain text directly extracted from web pages, significantly simplify downstream processing, particularly for tasks such as pre-training LLMs. A downside of WET's generic extraction is that it often includes all text on the web page, which could affect the quality of the datasets built from WET.

The differences between the two formats present a natural trade-off between the computational complexity and size of the data versus the quality and cleanliness of the extracted text. To provide perspective, the total size of WARC files accumulated from 2013 to 2024 was roughly 5.9 PB, while WET files in the same period were 751 TB, which is significantly smaller than their WARC counterparts. This substantial size difference underscores the computational costs associated with pursuing higher data quality through processing WARC files.

While processing WARC files may improve data quality through better extraction, the practical benefits remain uncertain, particularly for Arabic language data. Many people tend to use WET files instead of WARC, largely due to their reduced processing requirements. Nevertheless, given the potential data quality benefits of employing more robust extraction tools on WARC files, we decided to explore both file formats and evaluate the resulting datasets systematically.

\begin{figure}[t]
    \centering
    \includegraphics[width=0.90\textwidth]{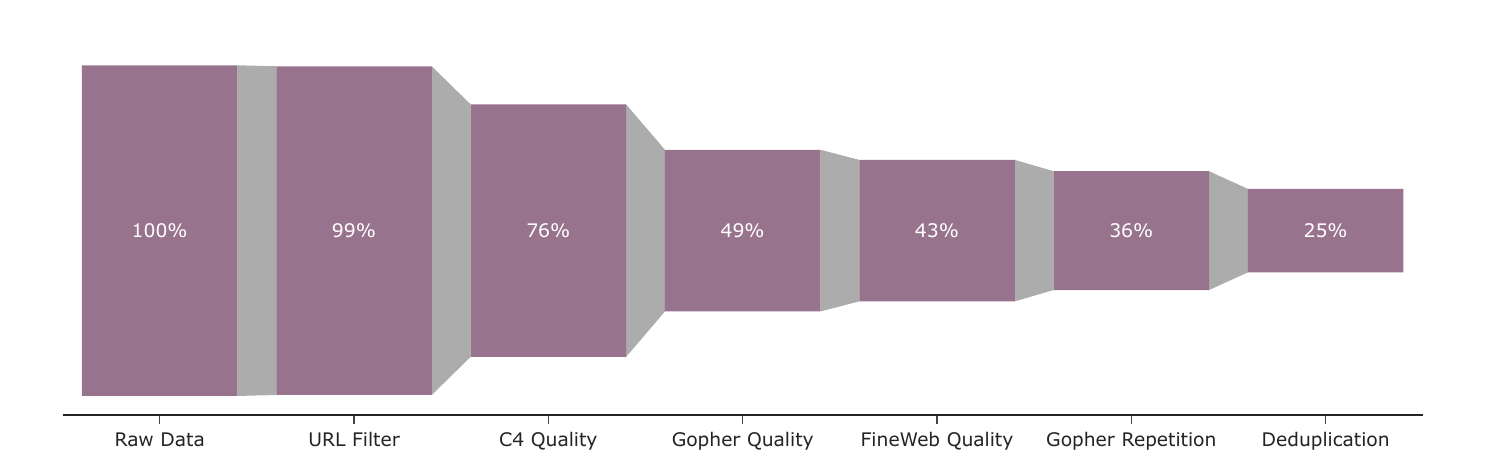}
    % \vspace{-0.15cm}
    \caption{The base filtration pipeline, composed of multiple filters in addition to deduplication (MinHash). Numbers indicate the percentage of web pages left after each filter.}
    \label{fig:base pipeline}
\end{figure}

\subsection{The Role of Pre-Training Data Quality}

Recent research has highlighted that while large datasets are essential, the quality of the data plays a more significant role in model performance than quantity.
The Colossal Clean Crawled Corpus (C4) demonstrated that straightforward language identification, boilerplate removal, and near-duplicate filtering can underpin strong models such as T5, underscoring the value of systematic cleaning \cite{dodge2021documentinglargewebtextcorpora}.
The Pile extended this idea by merging twenty-two curated sources and demonstrated that diversity, combined with deduplication, yields larger downstream gains than merely adding raw web text \cite{gao2020pile800gbdatasetdiverse}.
The Gopher project found that rigorous deduplication and removal of low-quality pages reduced perplexity even when the token budget was held constant \cite{rae2022scalinglanguagemodelsmethods}.
RedPajama-v2 builds on these lessons by attaching document-level quality scores and keeping only high-scoring pages, producing stronger models at the same budget \cite{together2023redpajama}.
FineWeb and its educational variant FineWeb-Edu advance quality control by adding lightweight model-based filters to their filtration pipeline, achieving consistent benchmark gains \cite{penedo2024finewebdatasetsdecantingweb}.
Finally, DataComp-LM supplies a 240-trillion-token pool and a benchmark that isolates data-curation effects, revealing that a strong model-based quality classifier is one of the most influential contributors to downstream performance gains. \cite{li2024datacomplmsearchgenerationtraining}.
Together these works underline that systematic curation, whether heuristic or model-based, is more influential than sheer token count and thus frames our study of high-quality Arabic pre-training data.

\subsubsection{Experimental Setup}
\label{sec:data_experimental_setup}

To evaluate the impact of Arabic data quality on pre-training performance, we conducted a systematic study focusing on different quality levels of Arabic corpora. The primary objective is to understand how varying data quality affects the performance of large language models. 
In the pre-training context, data quality refers to the dataset’s effectiveness in improving downstream task performance under fixed training conditions. All models were trained on the same number of tokens (25B) and under identical hyperparameters, ensuring that performance differences arise from data quality rather than dataset size or training setup. We therefore evaluate dataset quality through the accuracy of models trained on each dataset variant across multiple Arabic benchmarks.

To do so, and throughout this section, we pre-trained LLaMA3.2-1B \cite{llama3.2-1B} from scratch on various versions of the data representing varying levels of Arabic data quality and evaluated their performance on standard benchmarks.

The model consists of 1.23B parameters and was used with its associated tokenizer. Each model was trained on a random subset of approximately 25B tokens, which corresponds to the Chinchilla-optimal training size for this model scale \cite{chinchilla}. Training was conducted using the Llama-Factory \cite{llamafactory} framework for a single epoch with a sequence length of 2048 tokens, a batch size per device of 4, and an accumulation of gradients over 4 steps. The minimum learning rate was set to $5 \times 10^{-5}$, and the AdamW optimizer was used. To maintain computational efficiency, the model was trained using bf16 precision and on a hardware setup comprising 16 NVIDIA A100 GPUs.

Following prior work \cite{penedo2025fineweb2}, we evaluated the pre-trained Arabic models on a diverse set of datasets using the \textbf{LightEval} framework \cite{lighteval}. We selected 10 tasks specifically designed or adapted for Arabic from the \textbf{FineTasks} benchmark collection suggested by HuggingFace \cite{kydlicek2024finetasksmultilingualtasks}. These tasks help assess general knowledge, reasoning, and natural language understanding capabilities in Arabic. The datasets used for evaluation were 

\begin{enumerate}
    \item \textbf{General Knowledge (GK):} Arabic-Exam \cite{hardalov-etal-2020-exams}, Culture-Arabic-MMLU\footnote{Culture-Arabic-MMLU is a renamed version of the Arabic-MMLU dataset \cite{koto-etal-2024-arabicmmlu}, introduced here to distinguish it from the translated Arabic-MMLU dataset used later in this paper.} \cite{koto-etal-2024-arabicmmlu}, Alghafa (ARC) and Alghafa (SCIQA) \cite{almazrouei-etal-2023-alghafa}.
    
    \item \textbf{Reasoning (RES):} XCODAH \cite{Chen2019CODAHAA}, AlGhafa (PIQA) \cite{almazrouei-etal-2023-alghafa}, and XCSQA \cite{Talmor2019commonsenseqaaq}.
    
    \item \textbf{Natural Language Understanding (NLU):} XNLI-2.0 \cite{upadhyay2023xnli20improvingxnli}, MLMM (HellaSwag) \cite{zellers2019hellaswag}, and XStoryCloze \cite{DBLP:journals/corr/abs-2112-10668}.
\end{enumerate}

Similar to previous work \cite{messmer2025enhancingmultilingualllmpretraining}, we report the average normalized accuracy across all tasks as a general indicator of Arabic model performance, and detailed results for each task are presented in the following sections.

\subsubsection{Proposed Extraction and Filtration Pipeline}
\label{sec:Filtration_Pipeline}

Filtering unwanted content is a crucial step in preparing a high-quality Arabic pre-training dataset, as such content offers no benefit to downstream tasks and may degrade model performance. Web pages often include irrelevant material, such as non-Arabic text, noisy or poorly extracted content, and inappropriate material, including explicit content.
To study the impact of data quality in pre-training Arabic LLMs, we pre-train various models on varying levels of data filtering, in which we can assess the effectiveness of progressive filtering. Broadly, the filtration pipeline can be divided into three categories: (\textbf{1}) a language filter to identify Arabic web pages, (\textbf{2}) base quality filters consisting of heuristic rules and deduplication, and (\textbf{3}) model-based filters that apply Arabic-specific models to further improve quality.
Figure~\ref{fig:overall pipeline} provides a high-level overview of the proposed pipeline, from data extraction to the final model-based filtering stage.

\paragraph{Text Extraction} 
\label{para:text_extraction}

The first step of constructing a dataset from the web is text extraction. As noted earlier, WET files already contain extracted text and are ready for use, whereas WARC files require a text extraction process. We therefore focus on the two extractors used with WARC files, which are Trafilatura \cite{barbaresi-2021-trafilatura}
and Resiliparse \cite{zellers2019hellaswagmachinereallyfinish} 
extractors. Both libraries are effective at extracting plain text but come with distinct trade-offs. Trafilatura produces cleaner extractions by focusing on the main content and removing boilerplate, meaning repetitive page elements such as navigation menus, side panels, and cookie banners.
However, its heuristics can misclassify real content, like short captions or code blocks, as noise and discard them, this can sometimes lead to incomplete text.
Additionally, it is significantly slower than Resiliparse.

In contrast, Resiliparse offers a balance between Trafilatura's extraction and the plaintext offered in WET files. It tends to include some boilerplate and fewer advertisements than what we can find in WET, but it is faster than Trafilatura in the extraction, making it a practical choice for large-scale processing with limited computational resources. We report in the Appendix, a randomly sampled web page extracted using these two frameworks, alongside the WET file extraction. We can see how Trafilatura's extraction is cleaner and more concise, whereas Resiliparse and WET versions contain boilerplate and noise. In this example, Resiliparse produced longer text than WET, although our manual review suggthe sts that WET extracts are usually longer overall.

\paragraph{Language Filter} 
\label{para:language_filter}

The second step in constructing the dataset is filtering for Arabic web pages from the billions collected by \cite{commoncrawl}, which provides a language record in the metadata of each web page, indicating up to three detected languages. This helps accelerate language filtering. However, this language record is only available for crawls starting from about mid-2018 onward, earlier crawls lack language records. We have the following two-step general approach to extract the Arabic data from all crawls. (\textbf{1}) For the crawls with a language record, we utilized these records to select web pages where Arabic is among the top three detected languages. We then applied a language detection model (Lingua \cite{lingua_py}) to verify the primary language of each page. This balances the trade-off between the high cost of running a language filter on every web page and the inaccuracy of relying solely on Common Crawl’s classification, which we found to include some false positives. (\textbf{2}) For crawls without language records, we first identify pages containing any of the 10 most frequent Arabic letters in the content, and only then we apply the Lingua model to confirm the primary language. Since Arabic is not a major language in the crawls, as discussed in Section~\ref{sec:Pre-training}, it only represents 0.6\% of the internet. Given this scarcity, filtering based on the presence of Arabic letters significantly speeds up the process by quickly discarding non-Arabic-script web pages.

\paragraph{Quality Filters} 
\label{para:quality_filter}

Web pages often include low-quality elements such as advertisements, boiler-plate, empty pages, and spam. Quality filters often use heuristic rules to filter out this noise and keep real content.
The base quality filtration pipeline used to remove low-quality content is adapted from the approach proposed in the FineWeb dataset \cite{penedo2024finewebdatasetsdecantingweb}, with a different arrangement of filters. Figure~\ref{fig:base pipeline} illustrates our base filtration pipeline, along with the data reduction observed at each stage.

The pipeline consists of several groups of filters, each serving a distinct purpose. (\textbf{1}) \textit{URL Filter}
excludes web pages based on a blacklist of URLs and keywords appearing in the URL. (\textbf{2}) \textit{C4 Quality} \cite{dodge2021documentinglargewebtextcorpora} a set of heuristic rules to filter out gibberish and boilerplate texts. (\textbf{3}) \textit{Gopher Quality} \cite{rae2022scalinglanguagemodelsmethods} applies a set of heuristic rules targeting low-quality and poorly extracted web pages.
(\textbf{4}) \textit{FineWeb Quality} \cite{penedo2024finewebdatasetsdecantingweb} applies further heuristics designed to detect list-like web pages and content with repetitive lines. 
(\textbf{5}) \textit{Gopher Repetition} \cite{rae2022scalinglanguagemodelsmethods} measures how much a document repeats the same n-gram spans,  discarding texts that repeat themselves excessively, a pattern typical of boilerplate and spam.
(\textbf{6}) \textit{Deduplication} A fuzzy hash-based deduplication technique (MinHash) is applied at the crawl level.
It identifies and removes near-duplicate documents, helping reduce redundancy and improve training efficiency by ensuring more diverse content. 

To evaluate the effectiveness of our base filtration pipeline, we conduct experiments to assess its impact on both data quality and model performance. Specifically, we aim to study how data quality affects Arabic LLMs performance, and evaluate the pipeline’s ability to produce clean and useful text. To this end, we extract data from four stages of the filtering process, each reflecting increasing levels of filtering applied to the raw data. By pre-training models on each subset and evaluating them as described in Section~\ref{sec:data_experimental_setup}, we can quantify the contribution of each filtering stage. Grouping filters into these stages lets us measure quality gains without the costly step of testing every individual filter.
The four stages are:
% \vspace{-0.15cm}
\begin{enumerate}

  \item  \label{item:raw_data} \textbf{Raw}: the raw data of Common Crawl immediately after extraction, with no filtering applied (Step 1 in Figure~\ref{fig:base pipeline}).
  
  \item \textbf{Partially-Filtered}: only half of the filtering pipeline is applied to the raw data, up to and including the C4-Quality Filter. This results in some residual noise and redundancy remaining in the dataset (Step 2 to 3 in Figure~\ref{fig:base pipeline}).
  
  \item \textbf{Fully-Filtered}: the complete base filtering pipeline is applied to the raw data, except the deduplication step (Step 4 to 6 in Figure~\ref{fig:base pipeline}).
  
  \item \textbf{Deduplicated}: the full filtering pipeline is applied, including the MinHash deduplication step, ensuring that the data is both filtered and free of duplicates (Step 6 in Figure~\ref{fig:base pipeline}). 

\end{enumerate}

\subsubsection{Experiments}

\paragraph{Impact of Text Extraction on Arabic Pre-training Data} 

\setlength{\tabcolsep}{10pt}
\begin{wraptable}{r}{0.5\textwidth} 
\centering
\resizebox{0.5\textwidth}{!}{
\begin{tabular}{c|>
{\centering\arraybackslash}m{1.5cm}|>{\centering\arraybackslash}m{1.5cm}|>{\centering\arraybackslash}m{1.5cm}|>{\centering\arraybackslash}m{1.5cm}}
\hline
\textbf{} & \textbf{Total time (sec)} & \textbf{Average time (sec)} & \textbf{Total No. of words} & \textbf{Average No. of words} \\ \hline\hline
Trafilatura & 853 & 0.085 & 3,831,326 & 427 \\
Resiliparse & 25 & 0.002 & 10,581,006 & 1,058 \\
\hline
\end{tabular}
}
% \vspace{-0.15cm}
\caption{Extraction time and words count statistics for Trafilatura and Resiliparse on 10,000 Arabic web pages.}
\label{tab:extraction_comparison}
\end{wraptable}

To quantify the impact of text extraction methods on Arabic LLM data, we conducted a comparative evaluation using the three text extraction variants: WET, WARC (Resiliparse), and WARC (Trafilatura). Table~\ref{tab:extraction_comparison} compares the extraction speed and output of Trafilatura and Resiliparse on a randomly sampled set of 10,000 Arabic web pages. We observe that Resiliparse is 33× faster than Trafilatura but produces 2.76× more words on average,  highlighting that Trafilatura extracts potentially much cleaner text despite being significantly slower.

To assess which text extraction variant produces higher-quality data for Arabic LLM pre-training, we pre-trained an Arabic LLM using each extracted dataset. Following the same experimental setup described earlier, we applied our base filtration pipeline to all variants. Table~\ref{tab:WETvsWARC_Evaluation} reports the average accuracy across the evaluation datasets. \newline
The results show that WARC (Trafilatura) achieves the highest overall performance, with an average accuracy of 34.03\%. WARC (Resiliparse) and WET follow with 33.60\% and 33.39\%, respectively. While WARC (Resiliparse) slightly outperforms WET, both lag behind WARC (Trafilatura), which consistently delivers better results across most tasks.
These findings indicate that the Trafilatura extractor yields higher-quality Arabic data for LLM pre-training compared to the other extraction methods. Despite its slower throughput, we prioritized text quality and therefore used Trafilatura as the default extraction method for all subsequent experiments.

\paragraph{Impact of Quality Filters on Arabic Pre-training Data}

To evaluate the effect of each stage in our filtration pipeline on Arabic data quality, we pre-trained four models on WARC data extracted with Trafilatura, the best-performing extractor from the previous experiment. Each model corresponds to a different filtration stage. Table~\ref{tab:trafilatura_models} presents the average accuracy. The results show that while successive filtration stages generally improve data quality, their impact on performance varies across tasks. Starting from the Raw data, the Partially-Filtered stage shows an improvement in accuracy (from 33.61\% to 34.00\%), indicating an initial enhancement in data quality. The Fully-Filtered stage shows a minor decrease in accuracy (33.65\%), which may be worth investigating in future work. However, after the Deduplicated stage, the accuracy increases notably to 34.03\%, indicating that removing duplicates contributes to higher data quality. These results highlight how the base filtration pipeline progressively improves Arabic data quality and positively impacts model performance

While aggressive filtering can inevitably remove some valuable content, this trade-off is intrinsic to large-scale pre-training data curation. At the scale of hundreds of billions of tokens, filtering must rely on heuristics and model-based approximations rather than deterministic selection. As a result, a small portion of useful text may be discarded; however, the net effect remains strongly positive, as substantially more low-quality and noisy content is eliminated. Moreover, some potentially informative text may appear in irregular or inconsistent layouts, which can affect the extraction process and lead to their removal by heuristic rules. Recent efforts have explored reformatting or rewriting such cases to recover useful content, but this direction lies beyond the scope of our current study.

\begin{table}[t]
\centering
\resizebox{\textwidth}{!}{
\begin{tabular}{c|>{\centering\arraybackslash}m{1.5cm}|>{\centering\arraybackslash}m{1.5cm}|>{\centering\arraybackslash}m{1.5cm}|>{\centering\arraybackslash}m{1.5cm}|>{\centering\arraybackslash}m{1.5cm}|>{\centering\arraybackslash}m{1.5cm}|>{\centering\arraybackslash}m{1.5cm}|>{\centering\arraybackslash}m{1.5cm}|>{\centering\arraybackslash}m{1.5cm}|>{\centering\arraybackslash}m{1.5cm}|>{\centering\arraybackslash}m{1.5cm}}
\hline
\textbf{Model} & \textbf{All} & \textbf{Alghafa (ARC: easy)} & \textbf{Alghafa (SCIQA)} & 
\makecell{\textbf{Arabic-}\\\textbf{Exam}} & \makecell{\textbf{Culture-}\\\textbf{Arabic-}\\\textbf{MMLU}}
 & \textbf{Alghafa (PIQA)} & \textbf{XCODAH} & \textbf{XCSQA} & \textbf{MLMM (HellaSwag)} & \textbf{XNLI2} & \textbf{XStory Cloze} \\
\hline\hline
WET & 33.39 & 31.68 & 59.79 & 26.38 & 32.57 & 54.45 & 26.33 & \textbf{23.60} & 29.05 & 56.90 & 52.81 \\
\textbf{WARC (Trafilatura)} & \textbf{34.03} & \textbf{33.04} & 60.30 & 28.11 & \textbf{33.04} & \textbf{55.10} & 27.66 & 22.90 & \textbf{30.21} & \textbf{59.03} & \textbf{53.48} \\
WARC (Resiliparse) & 33.60 & 30.80 & \textbf{61.01} & \textbf{28.16} & 32.63 & 53.00 & \textbf{29.00} & 22.80 & 28.30 & 56.90 & 52.80 \\
\hline
\end{tabular}
}
% \vspace{-0.15cm}
\caption{Evaluation of pre-training LLaMA3.2-1B model from scratch on 25B tokens using WET and WARC data.}
\label{tab:WETvsWARC_Evaluation}
\end{table}

\begin{table}[t]
\centering
\resizebox{\textwidth}{!}{
\begin{tabular}{c|>{\centering\arraybackslash}m{1.5cm}|>{\centering\arraybackslash}m{1.5cm}|>{\centering\arraybackslash}m{1.5cm}|>{\centering\arraybackslash}m{1.5cm}|>{\centering\arraybackslash}m{1.5cm}|>{\centering\arraybackslash}m{1.5cm}|>{\centering\arraybackslash}m{1.5cm}|>{\centering\arraybackslash}m{1.5cm}|>{\centering\arraybackslash}m{1.5cm}|>{\centering\arraybackslash}m{1.5cm}|>{\centering\arraybackslash}m{1.5cm}}
\hline
\textbf{Model} & \textbf{All} & \textbf{Alghafa (ARC: easy)} & \textbf{Alghafa (SCIQA)} & \makecell{\textbf{Arabic-}\\\textbf{Exam}} & \makecell{\textbf{Culture-}\\\textbf{Arabic-}\\\textbf{MMLU}}
 & \textbf{Alghafa (PIQA)} & \textbf{XCODAH} & \textbf{XCSQA} & \textbf{MLMM (HellaSwag)} & \textbf{XNLI2} & \textbf{XStory Cloze} \\
\hline\hline
Raw & 33.61 & 32.06 & 60.50 & \textbf{31.56} & 32.24 & 55.21 & 27.67 & 22.60 & 28.65 & 54.20 & 53.01 \\
Partially-Filtered & 34.00 & 32.70 & 60.40 & 28.57 & 32.98 & \textbf{56.03} & \textbf{28.67} & \textbf{23.20} & 29.50 & 56.10 & 53.41 \\
Fully-Filtered & 33.65 & 32.91 & \textbf{61.71} & 29.34 & 32.37 & 55.16 & 27.00 & 22.80 & 29.82 & 58.30 & \textbf{54.07} \\
\textbf{Deduplicated} & \textbf{34.03} & \textbf{33.04} & 60.30 & 28.11 & \textbf{33.04} & 55.10 & 27.66 & 22.90 & \textbf{30.21} & \textbf{59.03} & 53.48 \\
\hline
\end{tabular}
}
% \vspace{-0.15cm}
\caption{Evaluation of Pre-training LLaMA3.2-1B model from scratch on 25B tokens across different WARC (Trafilatura) pipeline stages.}
\label{tab:trafilatura_models}
\end{table}

\begin{table}[!t]
    \centering
    \small
    \renewcommand{\arraystretch}{1.2}
    \setlength{\tabcolsep}{3pt} 

    \begin{subtable}[t]{0.49\textwidth}
        \centering
        \renewcommand{\arraystretch}{1.5} 
        \begin{tabular}
        {c c c} 
            \hline
            \textbf{Classifier} & \textbf{High Quality} & \textbf{Low Quality} \\
            \hline\hline
            A  & Wiki & Deduplicated WET \\
            B  & Wiki, 101B, SANAD & Raw WET \\
            C  & Wiki, Fineweb2 & Raw WARC (Resiliparse) \\
            \hline
        \end{tabular}
        % \vspace{-0.15cm}
        \caption{This table provides an overview of three FastText classifiers, each trained on a different combination of datasets representing high-quality and low-quality content. The dataset combinations were selected to assess how variations in data composition influence classification performance. Classifier A uses Wikipedia and  WET data after the deduplication stage; Classifier B combines Wikipedia and multiple curated datasets with Raw WET data; and Classifier C contrasts Wikipedia and FineWeb2 with Raw WARC (Resiliparse) data.}

        \label{tab:model_configurations}
    \end{subtable}%
    \hfill
    \begin{subtable}[t]{0.43\textwidth}
        \centering
        \begin{tabular}{ >{\centering\arraybackslash}p{2cm} c c c c } 
            \hline
            \multirow{2}{*}{\textbf{Classifier}}  & \multicolumn{2}{c}{\textbf{WET Data}} & \multicolumn{2}{c}{\textbf{WARC Data}} \\
            & \textbf{Acc(\%)} & \textbf{F1(\%)} & \textbf{Acc(\%)} & \textbf{F1(\%)} \\
            \hline\hline
            A & 94.00 & 0 & 58.00 & 16.00 \\
            B & 92.00 & 0 & 70.00 & 54.55 \\
            C & 56.00 & 21.43 & 92.00 & 91.67 \\
            \hline
        \end{tabular}

        \caption{
     We evaluate the performance of three FastText classifiers on human-labeled examples from two document formats: WET and WARC (Trafilatura). Classifiers A and B achieve high accuracy, but 0 F1-score on WET, failing to identify any high-quality samples. Classifier C performs better on WET but still shows limited effectiveness. On WARC, Classifier A performs poorly, Classifier B shows moderate performance, and Classifier C achieves both high accuracy and F1-score.}

        \label{tab:model_performance}
    \end{subtable}
    
   \caption{Summary of the classifier training configurations and their corresponding performance on the Annotated Subset. The results illustrate how different dataset combinations influence classifier performance across various document formats.}

\end{table}

\begingroup
\subsubsection{Model-Based Filtering}
\label{sec:ModelBasedFiltering}

The base filtration pipeline consists of multiple stages aimed at gradually improving the quality of the Arabic pre-training data through various rule-based filters. To further enhance this pipeline, we introduce model-based filtering techniques that complement and extend beyond traditional rule-based methods. Although the impact of data quality has been explored in other languages, large-scale studies of model-based filtering for Arabic remain limited.

\paragraph{FastText Quality Classifier}

We employed a supervised \textit{FastText} classifier \citep{joulin2016bag} to perform binary classification, distinguishing between high-quality and low-quality text. We trained three classifiers using a combination of datasets categorized by quality. Raw and Deduplicated WET data, along with Raw WARC (Resiliparse) (see Figure~\ref{fig:text extraction}), were considered low-quality due to their lack of curation. \\ Wikipedia articles~\cite{wikipedia}, the 101 Billion Arabic Words Dataset~\cite{aloui2024101}, Fineweb2 \cite{penedo2025fineweb2} and SANAD~\cite{hermessi2023sanad} were treated as high-quality sources, as they are curated and exhibit greater linguistic consistency. These datasets were selected to provide a balanced mix of data qualities, with the goal of improving the classifier’s ability to distinguish effectively between high- and low-quality texts. We used the default hyperparameters provided by FastText, except for the maximum word n-gram length, which we increased from the default value of 1 to 3. A summary of the configurations used is provided in Table~\ref{tab:model_configurations}.

\newpage
\paragraph{Human Evaluation for Ground Truth Labeling}
\label{GroundTruth}

To evaluate the performance of the trained FastText classifiers, we needed a reliable ground truth for comparison. To address this, we conducted a manual evaluation in which human annotators labeled a subset of WARC (Trafilatura) and WET data. This Annotated Subset was then used as the reference ground truth.

The subset consisted of 100 samples, evenly split between WET and WARC (Trafilatura) after the deduplication step described in Figure~\ref{fig:base pipeline}. We excluded WARC (Rasliparse) to avoid overlap with Classifier C’s training data, where it was used as the low-quality class.Although WET was used to train Classifiers A and B, it remains highly variable and loosely structured, which makes it a good proxy for the noisy, unpredictable nature of real-world web data. In contrast, WARC (Rasliparse) is more consistent and structured compared to WET, so using it in both training and evaluation could bias the results by making the classifier appear more effective than it is on more diverse or less structured data.

To create this Annotated Subset, each human annotator independently assigned a binary label: 1 for high-quality samples, and 0 for low-quality samples. The labeling process was guided by the criteria outlined in the Appendix, which specified how to assess the clarity and overall quality of each text. The final labels were then determined using a majority voting strategy.

\suppressfloats[t]    
\noindent

    \begin{wraptable}[18]{r}{0.4\textwidth}
    \centering
    \resizebox{0.4\textwidth}{!}{
        \begin{tabular}{ >{\centering\arraybackslash}p{1.8cm} c @{\hspace{40pt}} c }
            \hline
            \multirow{2}{*}{\textbf{Classifier}} & \multicolumn{2}{c}{\textbf{Metrics}} \\
            & \textbf{Acc (\%)} & \textbf{F1 (\%)} \\
            \hline\hline
            A &  53.85 & 14.29 \\
            B &  65.38 & 50.00  \\
            C &  90.38 & 90.91  \\
            \hline
        \end{tabular}
    }
        \captionsetup{type=table}
        \captionof{table}{
        This table summarizes the performance of FastText classifiers on a balanced subset of human-labeled examples, where each model was evaluated using equal numbers of high- and low-quality samples to remove class imbalance effects. Classifier C clearly outperforms the others, achieving high accuracy and F1-score consistent with human annotations, while Classifier A performs poorly and B shows moderate results.
        }
        \label{tab:balanced_model_performance}
    \end{wraptable}

\paragraph{Classifier Performance Against Ground Truth}
To assess the classifiers’ performance, we evaluated their predictions on the Annotated Subset described above. This allowed us to measure alignment with human judgments. We report our results in Table~\ref{tab:model_performance}. Classifiers A and B achieved high Accuracy on the WET subset but had an F1-score of 0, indicating that neither identified any high-quality examples correctly. On the WARC subset, Classifier A performed poorly, while Classifier B achieved moderate results. In contrast, Classifier C exhibited the best performance on the WARC subset but performed less effectively on WET. 
To further validate these findings, we evaluated all three classifiers on a balanced split of the Annotated Subset, consisting of 52 examples equally split between high- and low-quality labels. As shown in  Table~\ref{tab:balanced_model_performance}, Classifier C outperforms the others by a wide margin, achieving both high accuracy and F1-score, indicating strong alignment with human annotations. Classifier A, by contrast, shows weak performance across both metrics, while Classifier B demonstrates moderate effectiveness, though still with a noticeable gap behind Classifier C.

\paragraph{Educational Classifier}
\label{para:educational_classifier}

Another model-based filtering approach involves training a classifier to assess the educational value of the content \cite{penedo2024finewebdatasetsdecantingweb}. While this technique has been adopted in several non-public datasets, FineWeb-Edu made an effort to open-source the experiment with this type of technique, which we tried to replicate on the Arabic data.
To generate training data, we used the Qwen2.5-72B model \cite{Qwen2.5-72B-Instruct} to annotate a sample of 100K web pages, with educational level scores ranging from 0 to 5, where 5 denotes highly educational content. The choice of Qwen2.5-72B was informed by its strong performance on the FineTasks leaderboard \cite{kydlicek2024finetasksmultilingualtasks}, made as part of FineWeb2 \cite{penedo2025fineweb2}.

We experimented with several models and found that BGE-M3 \cite{bge-m3} performed best after fine-tuning on the synthetically annotated dataset, with care for the imbalance of classes in the dataset. The fine-tuned model obtained a macro F1-score of 0.48, closely matching the performance reported for the FineWeb-Edu classifier (0.50 macro F1) while using only 100K annotated samples, approximately one-fifth of the original training data (450K). This indicates that the educational-value classification approach transfers effectively to Arabic text, even with substantially fewer examples.\\
For the annotation prompt, we used the original prompt used by FineWeb-Edu and an Arabic-translated version of it. After testing them on a sample of 10K web pages, we found that the English prompt made the model better at instruction following, compared to the Arabic version. In addition to the different behavior of the annotation, Arabic was more conservative and assigned lower scores than the English prompt did, see Figure~\ref{fig:Edu heatmap} and Figure~\ref{fig:Edu distribution}. \\
Based on these results, we went forward with the English prompt to annotate the 100K dataset. The model trained on this dataset assigns a score from 0 to 5 to each web page. To categorize the corpus, we label web pages with scores of 0 or 1 as low quality, and those with scores from 2 to 5 as high quality.

\begin{table}[t]
\centering
\resizebox{\textwidth}{!}{
\begin{tabular}{c|>{\centering\arraybackslash}m{1.5cm}|>{\centering\arraybackslash}m{1.5cm}|>{\centering\arraybackslash}m{1.5cm}|>{\centering\arraybackslash}m{1.5cm}|>{\centering\arraybackslash}m{1.5cm}|>{\centering\arraybackslash}m{1.5cm}|>{\centering\arraybackslash}m{1.5cm}|>{\centering\arraybackslash}m{1.5cm}|>{\centering\arraybackslash}m{1.5cm}|>{\centering\arraybackslash}m{1.5cm}|>{\centering\arraybackslash}m{1.5cm}}
\hline
\textbf{Model} & \textbf{All} & \textbf{Alghafa (ARC: easy)} & \textbf{Alghafa (SCIQA)}&\makecell{\textbf{Arabic-}\\\textbf{Exam}}
&
\makecell{\textbf{Culture-}\\\textbf{Arabic-}\\\textbf{MMLU}}
& \textbf{Alghafa (PIQA)} & \textbf{XCODAH} & \textbf{XCSQA} & \textbf{MLMM (HellaSwag)} & \textbf{XNLI2} & \textbf{XStory Cloze} \\
\hline\hline
Deduplicated & 34.03 & 33.04 & 60.30 & 28.11 & 33.04 & 55.10 & 27.66 & 22.90 & 30.21 & \textbf{59.03} & 53.48 \\ \hline
FastText & 34.29 & 34.18 & 59.90 & 30.69 & 33.02 & \textbf{55.65} & 27.33 & 23.10 & 30.11 & 57.61 & \textbf{55.26} \\
\textbf{Educational} & \textbf{34.70} & \textbf{34.39} & \textbf{61.61} & \textbf{30.81} & \textbf{33.54} & 54.67 & \textbf{27.67} & \textbf{23.70} & \textbf{30.41} & 55.98 & 54.78 \\
\hline
\end{tabular}
}
\caption{Evaluation of Pre-training LLaMA3.2-1B model from scratch on 25B tokens of classified WARC (Trafilatura) data using model-based filtering: FastText and Educational Classifiers.}
\label{tab:model_based_evaluation}
\end{table}

\begin{figure}[!t]
    \centering
    \begin{minipage}[t]{0.48\textwidth}
        \centering
        \includegraphics[width=\textwidth]{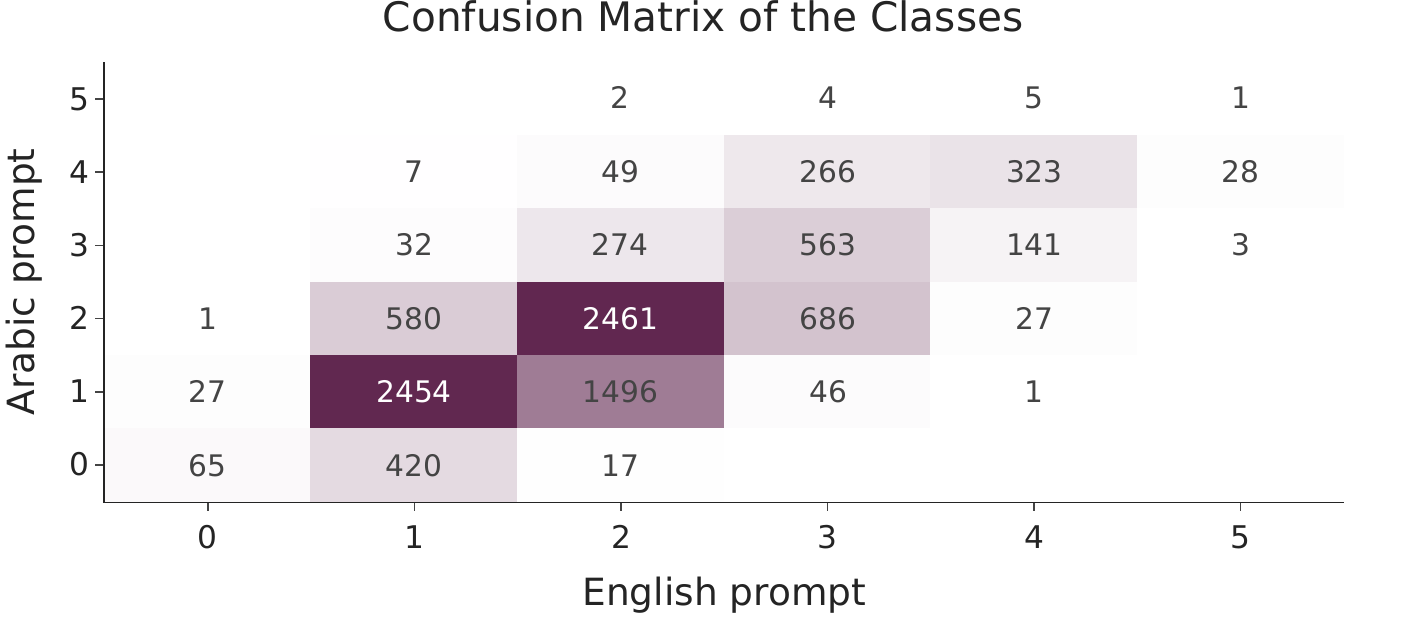}
        % \vspace{-0.5cm}
        \caption{The Confusion matrix of the classes assigned by the educational classifier for both English and Arabic prompts on a 10K sample. We can see the agreement on most of the web pages, but generally, the Arabic prompt tends to give lower scores compared to the English prompt.}
        \label{fig:Edu heatmap}
    \end{minipage}
    \hfill
    \begin{minipage}[t]{0.48\textwidth}
        \centering
        \includegraphics[width=\textwidth]{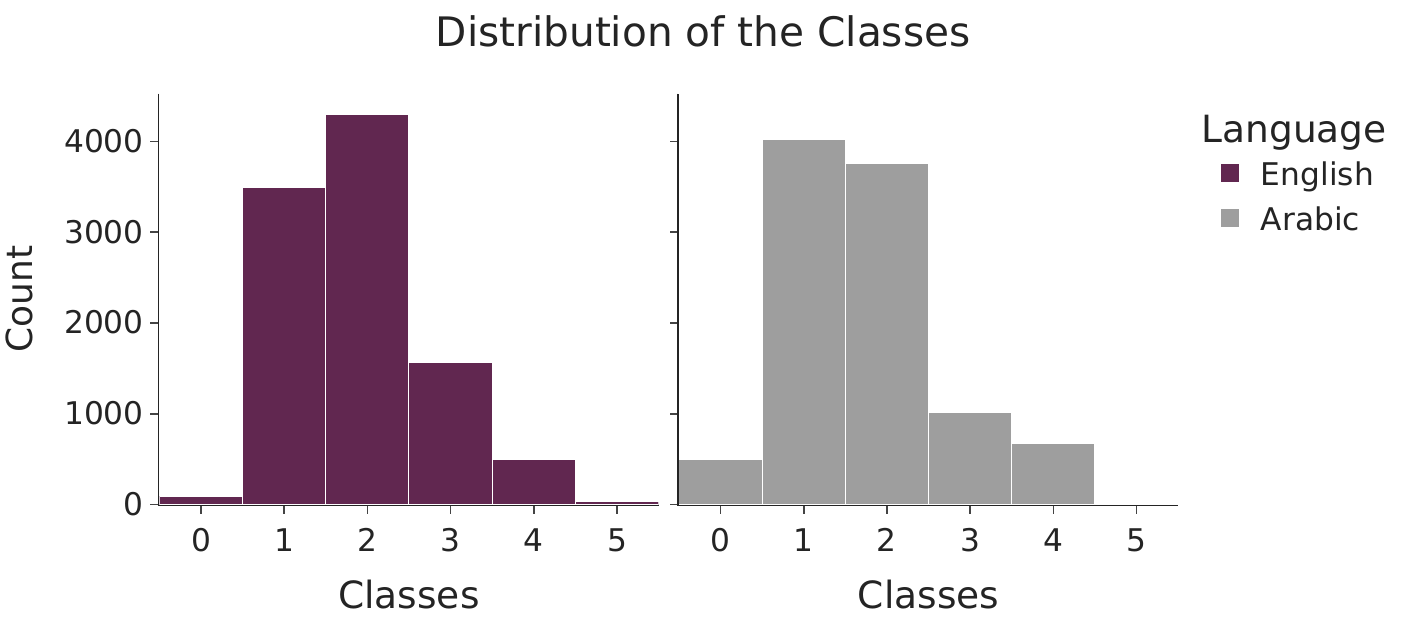}

        \caption{The distribution of the classes assigned by the educational classifier for both English and Arabic prompts on a 10K sample. While both are right-skewed, the Arabic prompt assigned more 0 and 1 classes compared to the English prompt}
        \label{fig:Edu distribution}
    \end{minipage}
    % \vspace{-0.8em}
\end{figure}

\paragraph{Impact of Model-Based Filtering on Arabic Pre-training Data} To assess the impact of model-based filtering on Arabic Pre-training data, we applied our two best-performing classifiers, FastText (Classifier C) and the Educational classifier, as an additional filtration layer on top of the base filtering pipeline using WARC (Trafilatura) data. A model was then pre-trained on the filtered data and evaluated across 10 benchmark tasks.

As shown in ~\autoref{tab:model_based_evaluation}, both model-based filtering classifiers outperformed the deduplication stage from the base filtration pipeline, confirming the added value of these quality classifiers. The ``FastText" model achieved an overall accuracy of 34.29\%, representing a +0.26 point improvement over the ``Deduplicated" model (34.03\%), while the "Educational" classifier reached 34.70\% with an enhancement of +0.67 points. 

We observed a trade-off between quality and quantity: FastText classifier retained more data (170.6B words), while the Educational classifier was more selective (136.9B words). Despite the smaller size, the Educational classifier achieved the highest accuracy, making it the strongest candidate for retaining high-quality Arabic data for LLM Pre-training.

\subsubsection{The Final Dataset: CuAra}

We present the final dataset \textbf{CuAra}, the end product of our complete filtration pipeline (see Figure~\ref{fig:overall pipeline}), which includes text extraction from WARC data using Trafilatura, followed by language and quality filtering, and finally model-based filtering using the Educational classifier.

We evaluated CuAra dataset against leading open-source Arabic datasets: \textbf{101B Arabic Words} \cite{aloui2024101}, \textbf{ArabicWeb24} \cite{ArabicWeb24}, and \textbf{FineWeb2} \cite{penedo2025fineweb2}. For each dataset, we pre-trained and evaluated three LLMs using randomly sampled subsets under the same experimental setup. We then evaluated the models and reported the average accuracy and standard deviation for each dataset. As shown in Table~\ref{tab:DataEvaluation}, the CuAra dataset achieved the highest overall accuracy (34.65\%) across different benchmark tasks. These results demonstrate the effectiveness of our filtration pipeline and the strong impact of model-based filtering (the Educational classifier) in producing high-quality Arabic data for LLM pre-training.

In terms of scale, CuAra datasets are significantly larger than existing open-source Arabic datasets. Our final datasets comprise \textbf{170.6B words} with the FastText classifier and \textbf{136.9B words} with the Educational classifier, compared to 22.6B words in 101B Arabic Words dataset, 17.7B words in ArabicWeb24, and 30.3B words in FineWeb2. This substantial increase in size provides a broader and more diverse foundation for Arabic LLM pre-training. 

% \newpage

\begin{table}[t]
\centering
\resizebox{\textwidth}{!}{
\begin{tabular}{c|
>{\centering\arraybackslash}m{1.5cm}|
>{\centering\arraybackslash}m{1.5cm}|
>{\centering\arraybackslash}m{1.5cm}|
>{\centering\arraybackslash}m{1.5cm}|
% Culture_Arabic MMLU column widened
>{\centering\arraybackslash}m{1.8cm}|
>{\centering\arraybackslash}m{1.5cm}|
>{\centering\arraybackslash}m{1.5cm}|
>{\centering\arraybackslash}m{1.5cm}|
>{\centering\arraybackslash}m{1.5cm}|
>{\centering\arraybackslash}m{1.5cm}|
>{\centering\arraybackslash}m{1.5cm}}
\hline
\textbf{Model} & \textbf{All} & \textbf{Alghafa (ARC: easy)} & \textbf{Alghafa (SCIQA)} &
\makecell{\textbf{Arabic-}\\\textbf{Exam}} &
\makecell{\textbf{Culture-}\\\textbf{Arabic-}\\\textbf{MMLU}}
 &
\textbf{Alghafa (PIQA)} & \textbf{XCODAH} & \textbf{XCSQA} & \textbf{MLMM (HellaSwag)} & \textbf{XNLI2} & \textbf{XStory Cloze} \\
\hline\hline
\textbf{CuAra (ours)}  &
\makebox[1.5cm][c]{\textbf{34.65 ± .35}} &
\makebox[1.5cm][c]{\textbf{34.40 ± .78}} &
\makebox[1.5cm][c]{\textbf{61.17 ± .67}} &
\makebox[1.5cm][c]{\textbf{29.84 ± .72}} &
\makebox[1.8cm][c]{\textbf{33.57 ± .34}} &
\makebox[1.5cm][c]{55.79 ± .58} &
\makebox[1.5cm][c]{27.33 ± 1.00} &
\makebox[1.5cm][c]{23.27 ± .35} &
\makebox[1.5cm][c]{30.44 ± .29} &
\makebox[1.5cm][c]{57.91 ± 1.58} &
\makebox[1.5cm][c]{53.85 ± .54} \\
101B Arabic Words &
\makebox[1.5cm][c]{31.54 ± .33} &
\makebox[1.5cm][c]{28.21 ± .10} &
\makebox[1.5cm][c]{57.92 ± 1.28} &
\makebox[1.5cm][c]{28.11 ± .52} &
\makebox[1.8cm][c]{30.40 ± .44} &
\makebox[1.5cm][c]{50.90 ± .55} &
\makebox[1.5cm][c]{26.45 ± .69} &
\makebox[1.5cm][c]{22.13 ± .29} &
\makebox[1.5cm][c]{26.59 ± .12} &
\makebox[1.5cm][c]{51.03 ± .61} &
\makebox[1.5cm][c]{51.89 ± 1.00} \\
ArabicWeb24 &
\makebox[1.5cm][c]{33.96 ± .28} &
\makebox[1.5cm][c]{34.26 ± .36} &
\makebox[1.5cm][c]{60.87 ± .65} &
\makebox[1.5cm][c]{28.20 ± 1.23} &
\makebox[1.8cm][c]{32.81 ± .32} &
\makebox[1.5cm][c]{\textbf{56.59 ± .47}} &
\makebox[1.5cm][c]{27.44 ± .84} &
\makebox[1.5cm][c]{\textbf{24.17 ± .45}} &
\makebox[1.5cm][c]{\textbf{30.49 ± .07}} &
\makebox[1.5cm][c]{\textbf{58.66 ± .82}} &
\makebox[1.5cm][c]{\textbf{53.98 ± .65}} \\
FineWeb2 (Arabic) &
\makebox[1.5cm][c]{33.52 ± .11} &
\makebox[1.5cm][c]{33.49 ± .39} &
\makebox[1.5cm][c]{60.47 ± .96} &
\makebox[1.5cm][c]{28.63 ± .04} &
\makebox[1.8cm][c]{32.27 ± .17} &
\makebox[1.5cm][c]{55.39 ± .73} &
\makebox[1.5cm][c]{\textbf{27.55 ± 1.07}} &
\makebox[1.5cm][c]{23.50 ± 1.45} &
\makebox[1.5cm][c]{30.19 ± .02} &
\makebox[1.5cm][c]{58.06 ± .24} &
\makebox[1.5cm][c]{53.92 ± .38} \\
\hline
\end{tabular}
}
% \vspace{-0.15cm}
\caption{Benchmark performance comparison of our data CuAra using the Educational classifier against Arabic baseline datasets: 101B Arabic words, ArabicWeb24, and FineWeb2. For each dataset, we pre-trained a LLaMA3.2-1B model from scratch on 25B tokens and evaluated performance across 10 tasks.}
\label{tab:DataEvaluation}
\end{table}

Beyond scale, CuAra differs from previous Arabic datasets in three main aspects. First, it is constructed directly from the raw WARC archives of Common Crawl rather than from pre-extracted WET files, enabling finer control over extraction quality and text structure. Second, its multi-stage filtering pipeline emphasizes quality at every level, combining heuristic filters, deduplication, and model-based filtering using Arabic-specific models such as FastText and the Educational classifier to reduce noise while preserving linguistic diversity. Together, these design choices result in a larger, cleaner, and more reliable dataset for Arabic LLM pre-training.

\begin{tcolorbox}[colback=black!5!white,colframe=black!75!black,title=Conclusion]
This work addresses the significant challenge of developing high-quality Arabic pre-training datasets essential for advancing large language models for the Arabic language. We employed a rigorous multi-stage pipeline, starting with large-scale collection from Common Crawl, followed by heuristic-based rules, and model-based filtering techniques. Our detailed ablation studies quantified the effectiveness of each stage, demonstrating the critical impact of careful dataset preparation on the performance of Arabic LLMs. This work underscores the importance of careful dataset curation in overcoming current limitations and establishing a solid foundation for robust and capable Arabic LLMs.
\end{tcolorbox}
\endgroup
\section{Tokenization}
\noindent Recent research highlights the critical impact of tokenizer configurations on LLM performance across tasks and languages. \cite{ali2024tokenizer} demonstrated that factors like algorithm choice (Byte Pair Encoding (BPE) and Unigram), libraries implementing the training of tokenizers (Huggingface \cite{Moi_HuggingFace_s_Tokenizers_2023} and  SentencePiece \cite{SentencePiece}), and vocabulary size significantly influence outcomes. 

\cite{ahia2023languages,petrov2023language} investigates tokenization efficiency in multilingual models, revealing that languages like Arabic require significantly more tokens per sentence compared to Latin-based languages, leading to inefficiencies in processing. These higher tokenization costs, due to increased token counts, degrade model performance, especially in low-resource languages. This highlights the need for language-specific optimizations, particularly for complex, non-Latin scripts like Arabic, to improve compression and downstream performance. \cite{dagan2024tokenizer} showed that domain-specific tokenizers (e.g., code) improve compression and inference speed but introduce trade-offs in decoding costs. However, gaps remain in evaluating tokenizer effects across diverse contexts and balancing efficiency, cost, and performance.

While extensive research has explored various design choices for English, such as vocabulary size, training data composition, and pre-tokenization methods, similar studies for Arabic remain sparse and underexplored. Arabic’s unique linguistic characteristics, including its morphology, script, and dialectal variations, pose distinct challenges that require further investigation.

In this work, we address this gap by systematically benchmarking how various tokenizer design choices impact Arabic LLM performance. We evaluate a range of design choices to identify the most effective strategies for Arabic. Specifically, we trained multiple tokenizers with three distinct vocabulary sizes (32K, 64K, and 128K) and assessed the impact of various pre-tokenization methods. For the 64K vocabulary, we further explored different pre-tokenization approaches, including whitespace, ByteLevel \cite{huggingface_bytelevel}, GPT-4 Split ByteLevel \cite{openai2023gpt4}, and Punctuation Split ByteLevel\cite{dagan2024tokenizer}. Through these empirical evaluations, we provide actionable recommendations to optimize tokenization for Arabic LLMs, thereby enhancing both their efficiency and overall performance.

\subsection{Experiments Setup} \label{exp-setup}

We outline the experimental setup used to benchmark and evaluate various design choices for Arabic tokenization. Specifically, we focus on examining the impact of vocabulary size, pre-tokenization methods, and training data composition. We describe the dataset construction, tokenizer training, pre-training methodology, and evaluation framework. These components ensure a comprehensive assessment of the tokenization design choices and their influence on model performance across different settings.

\paragraph{Dataset:} We benchmark our results on a comprehensive dataset sourced from multiple domains, which we call \textbf{TokenMain}. This diverse dataset covers Arabic, English, mathematical content, and programming code, providing a broad basis for evaluating tokenizer design choices. The dataset is broken down as follows:

\begin{enumerate}
    \item \textbf{CC100 \cite{wenzek-etal-2020-cc100}:} Subsets from the Arabic and English portions of the multilingual CC100 corpus.
 \item \textbf{Wikipedia (wiki) \cite{wikidump}:} A subset of articles from both English and Arabic Wikipedia.
  \item \textbf{United Nations (UN) \cite{ziemski-etal-2016-un_ar_en}:}Official documents from the United Nations in both English and Arabic.
   \item \textbf{Math \cite{hendrycksmath2021}:} The MATH dataset, featuring challenging mathematics problems and solutions.
    \item \textbf{Code \cite{codeparrot2021}:} A subset of the GitHub Code dataset, sourced from open-source repositories.
    \item \textbf{Watan \& Khaleej (W\&K):} The Khaleej-2004 \cite{khaleej} corpus with 5,000 articles on news topics, and the Watan-2004 \cite{Watan} corpus with 20,000 articles across various topics.
    \item \textbf{Shamela Corpus \cite{shamela}:} A historical Arabic corpus containing classical texts, including works on Islamic theology and literature.
\end{enumerate}
 
We constructed specific data subsets from \textbf{TokenMain} to support two purposes: (i) \textbf{model pre-training} and (ii) \textbf{tokenizer training}, in which data used for \textbf{model pre-training} did not overlap with the subsets reserved for \textbf{tokenizer training}. The breakdown of these subsets is provided in Table~\ref{table:data_distribution}.
For \textbf{model pre-training}, we used a subset comprising 90\% of TokenMain, while preserving the original distribution across all subsets.  
For \textbf{tokenizer training}, we constructed the following subsets:
\begin{enumerate}
   \item \textbf{Heldout}: A subset comprising 5\% of the TokenMain, extracted while maintaining the original distribution across all subsets. 
    \item \textbf{ArWiki Heldout}: This subset also comprises 5\% of TokenMain while maintaining the original distribution for English, codes, and math, but all the Arabic subsets are from ArWiki \cite{wikidump}.

    \item \textbf{Reweighted Heldout}: This subset contains 5\% of the TokenMain, maintaining the original distribution for English, code, and math. However, for the Arabic subset, the distribution was adjusted to be proportional to $ \nicefrac{1}{p_i}$ where $p_i$ refers to the original distribution of Arabic data in TokenMain. 
\end{enumerate}

\paragraph{Tokenizer training framework: } For tokenizer training, we use the Hugging Face Tokenizers library \cite{Moi_HuggingFace_s_Tokenizers_2023} throughout.

\paragraph{LLM pre-training:}For pre-training LLM, we used the LLaMA-Factory framework \cite{llamafactory} with 8 NVIDIA A100 GPUs and a batch size of 64. A cosine learning rate scheduler was used, starting with an initial learning rate of 5$\times 10^{-5}$. The training process was executed for one epoch, and to maintain computational efficiency, we use 16 bits precision for training.

% \subsubsection{Evaluation}
\paragraph{Evaluation:}All models throughout the experiments were fine-tuned on a randomly selected 10\% subset of the Arabic benchmark datasets, including ACVA \cite{ACVA}, Alghafa \cite{almazrouei-etal-2023-alghafa}, Culture-Arabic-MMLU \cite{koto-etal-2024-arabicmmlu}, and Arabic-Exam \cite{hardalov-etal-2020-exams}, as shown in Table \ref{table:fine_tuning_data}. The remaining 90\% of the data was reserved for evaluation to ensure a fair assessment of model performance. All evaluations were conducted using the LightEval \cite{lighteval} framework.

\begin{figure}[t]
\noindent
\begin{minipage}[t]{0.49\textwidth}
\centering
\renewcommand{\arraystretch}{1.2}
\setlength{\tabcolsep}{4pt}
\resizebox{\textwidth}{!}{
\begin{tabular}{c|c|c|c|c|c}
\hline
\textbf{Subset} & \multicolumn{5}{c}{\textbf{Percentage}} \\ \cline{2-6}
 & \textbf{TokenMain} & \textbf{Model Pre-training} & \textbf{Heldout} & \textbf{Rew. Heldout} & \textbf{ArWiki Heldout} \\
\hline \hline
cc100     & 47.97 & 49.23 & 47.97 & 25.31 & - \\ \hline
en\_wiki  & 40.88 & 40.88 & 40.88 & 40.88 & 40.88 \\ \hline
ar\_wiki  & 7.37  & 6.38  & 7.37  & 25.48 & 56.24 \\ \hline
codes     & 1.30  & 1.30  & 1.30  & 1.30  & 1.30 \\ \hline
un\_en    & 0.91  & 0.91  & 0.91  & 0.91  & 0.91 \\ \hline
un\_ar    & 0.71  & 0.63  & 0.71  & 2.06  & - \\ \hline
math      & 0.65  & 0.65  & 0.65  & 0.65  & 0.65 \\ \hline
w\&k      & 0.14  & -     & 0.14  & 2.73  & - \\ \hline
shamela   & 0.03  & -     & 0.03  & 0.63  & - \\
\hline
\end{tabular}
}
\vspace{-0.15cm}
\captionsetup{type=table}
\captionof{table}{Percentage distribution of each data subset across phases (TokenMain, Model Pre-training, and Heldout sets).}
\label{table:data_distribution}
\end{minipage}%
\hfill
\begin{minipage}[t]{0.49\textwidth}
\centering
\renewcommand{\arraystretch}{1.2}
\resizebox{\textwidth}{!}{
\begin{tabular}{c|c|c}
\hline
\textbf{Model} & \textbf{Train Size} & \textbf{Question Type} \\
\hline \hline
ACVA           & 811  & True/False \\
Arabic-Exam    & 51   & MCQ (4 choices) \\
Culture-Arabic-MMLU    & 1327 & MCQ (4 choices) \\
Alghafa  & 2178 & MCQ (2–5 choices) \\
\hline
\end{tabular}
}
\captionsetup{type=table}
\captionof{table}{Summary of fine-tuning datasets used in the experiments, including dataset size and question formats.}
\label{table:fine_tuning_data}
\end{minipage}
\end{figure}

\subsection{Effect of Vocabulary Size}
Vocabulary size \cite{vocabsize}, defined as the number of unique tokens (such as words, sub-words, or characters) a tokenizer can recognize, is a crucial factor influencing tokenization efficiency and the performance of LLMs.
One key metric for evaluating tokenization efficiency is the fertility score \cite{fertilitybert, rust-etal-2021-good}, which measures the average number of tokens needed to represent a given text segment. A high fertility score indicates that more tokens are required for the same segment of text, suggesting inefficiency, while a low fertility score represents a more efficient tokenization process, with fewer tokens needed for the same text.

\begin{figure}[t]
    \centering
    % First graph
    \begin{subfigure}[b]{0.45\textwidth}
        \centering
        \includegraphics[width=\textwidth]{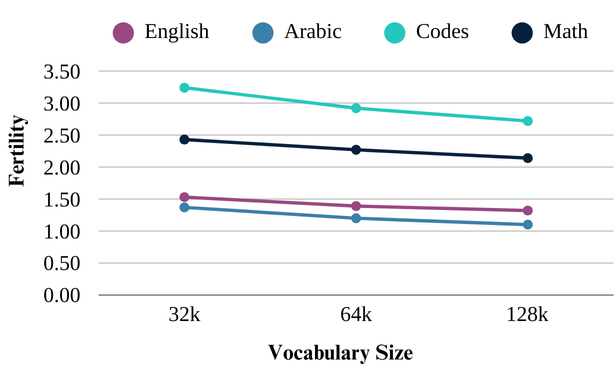}
        \label{fig:graph1-voc-fer}
    \end{subfigure}
    \hfill
    % Second graph
    \begin{subfigure}[b]{0.45\textwidth}
        \centering
        \includegraphics[width=\textwidth]{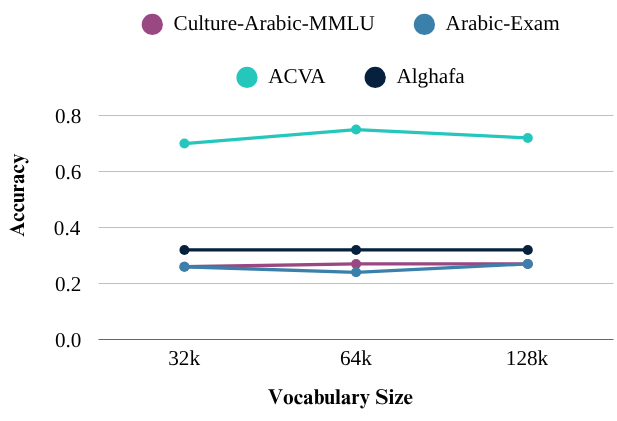}
        \label{fig:graph2-voc-acc}
    \end{subfigure}
    \vspace{-0.15cm}
    \caption{Comparison of fertility scores and vocabulary sizes (left) and downstream Arabic dataset accuracy and vocabulary sizes (right) for the pre-trained LLaMA3.2-1B model with a tokenizers trained on Arabic text. Fertility scores generally decrease with larger vocabularies; however, this does not consistently result in higher accuracy across all evaluated datasets}
    \label{fig:vocab-size-fer-acc}
\end{figure}

To investigate the effect of vocabulary size on tokenizer efficiency, fertility scores were computed using a subset of the model pre-training data that was not part of the tokenizer's training datasets: \textbf{Heldout}, \textbf{ArWiki Heldout}, and \textbf{Reweighted Heldout}. Figure~\ref{fig:vocab-size-fer-acc} (left) shows that fertility consistently decreases as vocabulary size increases from 32K to 128K across all domains. For English, fertility drops from 1.53 to 1.32. Arabic, a morphologically rich language, exhibits a sharper decrease from 1.37 to 1.10, indicating that larger vocabularies are more effective at efficiency and compression, even in morphologically complex contexts.

Structured domains such as Math and Code, however, maintain relatively high fertility scores even with larger vocabularies. Math decreases only to 2.14, while Code remains above 2.70 across all vocabulary sizes. This indicates that increasing vocabulary size significantly improves tokenization efficiency in morphologically rich languages such as Arabic, but yields only marginal gains in symbol-heavy domains, including Math and Code.

We further examined whether reduced fertility correlates with the LLM performance on downstream tasks, i.e., on the Arabic evaluation datasets. Using the LLaMA3.2-1B model, we replaced the default tokenizer with our \textbf{Heldout} tokenizer and then pre-trained the model with the modified tokenizer to further explore the impact of different vocabulary sizes. As shown in Figure~\ref{fig:vocab-size-fer-acc} (right), while fertility scores declined with larger vocabularies, indicating more efficient encoding, this did not consistently lead to higher accuracy on Arabic tasks across all evaluation datasets. These findings suggest that while increasing vocabulary size enhances tokenization efficiency, especially in morphologically rich languages, such gains do not necessarily translate into improved downstream model performance.

%#________________________#

\subsection{Effect of Pre-tokenization Methods}
Pre-tokenization \cite{dagan2024tokenizer} is a preprocessing step where text is split into smaller units, like words or punctuation, before the main tokenization process. It typically uses simple rules, such as spaces and punctuation, to create clear token boundaries, ensuring the tokens are meaningful for further analysis.

\begin{figure}[t]
    \centering
    % First graph
    \begin{subfigure}[b]{0.45\textwidth}
        \centering
        \includegraphics[width=\textwidth]{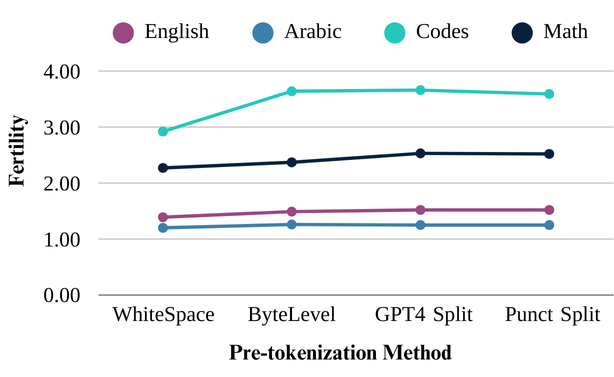}
        % \caption{Fertility scores across different pre-tokenization methods.}
        \label{fig:graph1}
    \end{subfigure}
    \hfill
    % Second graph
    \begin{subfigure}[b]{0.45\textwidth}
        \centering
        \includegraphics[width=\textwidth]{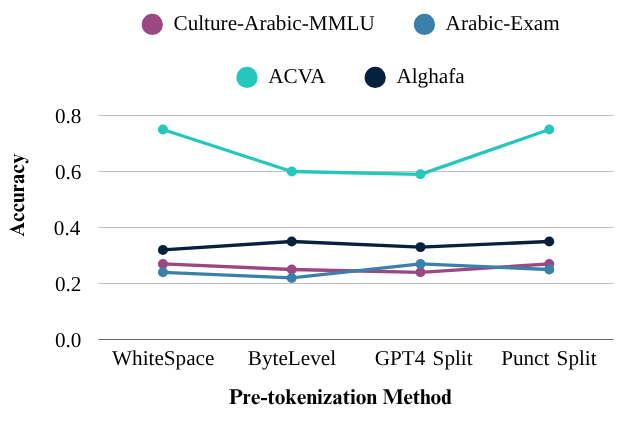}
        % \caption{Accuracy scores across different pre-tokenization methods.}
        \label{fig:graph2}
    \end{subfigure}
    \vspace{-0.15cm}
    \caption{Comparison of fertility scores and pre-tokenization methods (left) and downstream Arabic dataset accuracy and pre-tokenization methods (right) for the pre-trained LLaMA3.2-1B model with a tokenizers trained on Arabic text. Pre-tokenization methods have only a minor impact on both tokenization efficiency and downstream accuracy.}
    \label{fig:pre-tokeniz-fer-acc}
\end{figure}

As shown in Figures~\ref{fig:pre-tokeniz-fer-acc}, pre-tokenization methods have only a minor impact on both tokenization efficiency and downstream accuracy. ByteLevel approaches yield slightly higher fertility scores than whitespace. For example, in English, fertility increases from 1.39 (whitespace) to 1.49 (ByteLevel), and up to 1.52 with the GPT-4-style split, suggesting limited efficiency trade-offs. Structured domains like Math (2.52) and Code (3.59) show slightly higher fertility with Punctuation-split ByteLevel; however, overall differences remain small.

The effect of pre-tokenization on downstream accuracy varies across tasks. For Culture-Arabic-MMLU and Arabic-Exam, whitespace achieves the highest performance (0.27 and 0.24, respectively), whereas ByteLevel and GPT-4 Split perform slightly worse. Conversely, in ACVA and Alghafa, Punctuation Split matches or slightly surpasses Whitespace (0.75 and 0.35, respectively). These results indicate that while pre-tokenization methods can slightly influence fertility and downstream performance, no single strategy consistently outperforms others across domains and tasks.

%#________________________#
\subsection{Effect of Tokenizer Data Distribution on Downstream Performance }
To investigate the role of tokenizer data distribution and whether aligning it with the model's fine-tuning data distribution improves downstream performance, we conducted a series of experiments comparing different tokenizer configurations.

We constructed several tokenizers variants, each trained on a different subset of the Arabic corpus, designed to reflect distinct data distributions: \textbf{Heldout}, \textbf{ArWiki Heldout}, and \textbf{Reweighted Heldout}. These subsets, detailed in Section~\ref{exp-setup}, were designed to isolate the impact of distributional alignment between the tokenizer training data and the model fine-tuning data.

All tokenizers were trained with a fixed vocabulary size of 128K and applied whitespace pre-tokenization. Each tokenizer was then used to replace the default tokenizer of the LLaMA3.2-1B model during continued pre-training on Arabic data.

Table~\ref{tab:llama_different_tokenizers} reports the performance of LLaMA3.2-1B on various Arabic evaluation datasets when paired with different tokenizers trained on different datasets. We observe that tokenizers trained on data distributions more closely aligned with the fine-tuning data consistently yield better performance, with those trained on such subsets outperforming the default LLaMA3.2-1B tokenizer across all evaluation datasets.

For example, using the default LLaMA3.2-1B model and tokenizer without any pre-training on Arabic datasets, we achieve moderate performance across all tasks (0.60 accuracy on ACVA and 0.23 on Culture-Arabic-MMLU). Continued pre-training on Arabic data while keeping the default tokenizer fixed improves these scores to 0.70 and 0.24, respectively. Replacing the default tokenizer with the \textbf{Heldout} tokenizer, which is trained on the Heldout data subset, and applying the same continued Arabic pre-training further increases ACVA accuracy to 0.72 and Culture-Arabic-MMLU to 0.27, demonstrating the benefit of distributional alignment.

These results highlight that tailoring tokenizers to downstream data distributions can enhance tokenization efficiency and model performance. However, the optimal degree of alignment between tokenizer training and fine-tuning data remains an open question, across all experiments, including those involving the \textbf{ArWiki Heldout} and \textbf{Reweighted Heldout} datasets, variations in tokenization data still resulted in distributions that were more aligned with the fine-tuning data. Furthermore, models trained with customized tokenizers that are closely aligned with the downstream task distributions consistently outperformed those relying on the default LLaMA3.2-1B tokenizer.

\begin{table}[t]
\centering
\resizebox{\textwidth}{!}{%
\begin{tabular}{c|c|c|c|cccc}
\hline
\textbf{Model} &
  \textbf{Tokenizer} &
  \textbf{Vocab Size} &
  \textbf{Pre-trained} &
  \multicolumn{4}{c}{\textbf{Accuracy}} \\ \hline\hline
 &
   &
   &
   &
  \multicolumn{1}{c|}{Culture-Arabic-MMLU} &
  \multicolumn{1}{c|}{Arabic-Exam} &
  \multicolumn{1}{c|}{ACVA} &
   \multicolumn{1}{c}{Alghafa} \\ \hline
\multirow{3}{*}{LLaMA3.2-1B} &
  \multirow{2}{*}{LLaMA3.2-1B} &
  \multirow{2}{*}{128k} &
  $\times$ &
  \multicolumn{1}{c|}{0.23} &
  \multicolumn{1}{c|}{0.24} &
  \multicolumn{1}{c|}{0.60} &
  0.31 \\ \cline{4-8}
 &
  &
  &
   \checkmark &
  \multicolumn{1}{c|}{0.24} &
  \multicolumn{1}{c|}{0.26} &
  \multicolumn{1}{c|}{0.70} &
  0.32 \\ \cline{2-8}
 &
  Heldout Tokenizer &
  128k &
   \checkmark &
  \multicolumn{1}{c|}{\textbf{0.27}} &
  \multicolumn{1}{c|}{\textbf{0.27}} &
  \multicolumn{1}{c|}{0.72} &
  0.32 \\\hline\hline
  \multirow{2}{*}{LLaMA3.2-1B}
 &
  Arwiki Heldout Tokenizer &
 \multirow{2}{*}{128k} &
  \multirow{2}{*}{ \checkmark} &
  \multicolumn{1}{c|}{0.26} &
  \multicolumn{1}{c|}{0.25} &
  \multicolumn{1}{c|}{\textbf{0.75}} &
 \textbf{ 0.34 }\\ \cline{2-2}\cline{5-8}
 &
  Reweighted Heldout Tokenizer &
  &
   &
  \multicolumn{1}{c|}{\textbf{0.27}} &
  \multicolumn{1}{c|}{\textbf{0.27}} &
  \multicolumn{1}{c|}{0.72} &
  0.33 \\ \hline
\end{tabular}
}
\vspace{-0.15cm}
\caption{Accuracy comparison of LLaMA3.2-1B models pre-trained with various tokenizer configurations. Tokenizers differ in the distribution of Arabic data used for training. Models using tokenizers better aligned with the pre-training corpus show improved downstream performance.}
\label{tab:llama_different_tokenizers}
\end{table}

\begin{tcolorbox}[colback=black!5!white,colframe=black!75!black,title=Conclusion]
We explored how various Arabic tokenizer design choices affect Arabic LLM performance. Our findings show that (\textbf{i}) Increasing vocabulary size improves tokenization efficiency, particularly for Arabic, though this doesn't always lead to higher accuracy across tasks; (\textbf{ii}) Pre-tokenization for Arabic data does not seem to impact downstream performance; (\textbf{iii}) Tokenizers aligned with the fine-tuning data distribution  consistently deliver higher performance, highlighting the importance of data alignment.
\end{tcolorbox}

\section{Evaluation}

In this section, we focus on evaluating LLMs in Arabic, a language with unique structural and cultural characteristics that impact model performance. We review existing datasets in Section~\ref{sec:datasets} that evaluate LLM capabilities, particularly in Arabic, and identify key strengths and limitations. While English benchmarks are numerous and well established, Arabic benchmarks are fewer and often derived from translated datasets, which introduces issues related to cultural alignment and task fidelity. In Section~\ref{sec:leaderboard}, we introduce a comprehensive leaderboard that assesses Arabic and multilingual models on enhanced evaluation datasets, including those aligned with cultural contexts such as Saudi Arabia. The results reveal persistent challenges in Arabic language model evaluation compared to English, particularly with complex tasks and translated benchmarks.Together, these sections underscore the need for culturally relevant and diverse benchmarks to advance fair and effective LLM evaluation.

\subsection{Challenges in MMLU: A Focus on English, Arabic, and New Saudi Culture Dataset}\label{sec:datasets}

While English benchmarks are abundant and diverse, the evaluation landscape for Arabic remains relatively limited. Furthermore, even English datasets struggle to fully capture complex reasoning and real world knowledge challenges that often manifest, and in some cases intensify, in Arabic.

\noindent \textbf{English Evaluation Datasets.} 
English benchmarks cover various tasks aimed at evaluating general knowledge across domains (e.g., science, history, literature) and core language understanding skills such as reasoning, inference, and classification. For instance, 
 General Language Understanding Evaluation (GLUE) \cite{wang2019gluemultitaskbenchmarkanalysis} evaluates core natural language understanding (NLU) capabilities across nine tasks but is limited to single sentence or sentence pair evaluations, restricting its ability to assess complex language structures. To address this, SWAG \cite{zellers2018swagaf} and HellaSWAG \cite{zellers2019hellaswagmachinereallyfinish} focus on commonsense reasoning, though they are sensitive to linguistic ambiguity. The limitations of individual benchmarks led to multitask frameworks like BIG-Bench \cite{srivastava2023imitationgamequantifyingextrapolating}, which evaluates models across tasks like question answering, summarization, and translation.

\noindent \textbf{Arabic Evaluation Datasets.} 
Similarly, Arabic benchmarks aim to evaluate core capabilities such as reasoning, general knowledge, and language understanding, but they face unique challenges stemming from linguistic diversity and cultural specificity. The ARB-MMLU \cite{mmlu_arabic} adapted from its English counterpart (ENG-MMLU)~\cite{mmlu}, includes 15,908 multiple-choice questions drawn from North African, Gulf, and Levant curricula. While it provides high quality native Arabic content, its focus on Modern Standard Arabic (MSA) limits its applicability to dialectal variants. For instance, Alghafa ~\cite{almazrouei-etal-2023-alghafa} assesses zero-shot and few-shot performance but relies heavily on machine translated tasks, which can introduce semantic distortion, reduce cultural relevance, and misrepresent natural Arabic usage. ACVA~\cite{ACVA} takes a culturally grounded approach by evaluating models on true/false questions across 58 domains, though its binary format lacks the depth needed to test complex reasoning.

A persistent trend in Arabic LLM evaluation is the heavy reliance on direct translations of English benchmarks, see Table~\ref{tab:english_arabic_benchmarks}. While this approach expands the quantity of available Arabic datasets, it introduces two key problems. First, translated benchmarks often fail to reflect the full cultural and linguistic diversity of Arabic, limiting their ability to capture region specific nuances and contextually relevant tasks. Second, the process of translation itself can introduce additional errors and ambiguities. These issues are compounded by the fact that many original English datasets are not error free containing labeling mistakes, inconsistencies, or ambiguous phrasing, as seen in resources like CoNLL-2003\cite{CoNLL-2003} and ANERcorp\cite{ANERsys}, which required later corrections\cite{pang-etal-2020-conll, CLEANANERCorp}. When such flawed datasets are translated, their original errors are not only preserved but can be further exacerbated, ultimately undermining the accuracy and reliability of Arabic model evaluation.

\begin{table*}[t]
\centering
\scriptsize
\caption{Examples of English Benchmarks Translated into Arabic for LLM Evaluation. \textit{*Arabic\_Exam is not a translation of EXAMS but an Arabic subset for cross-/multilingual QA.}}
\label{tab:english_arabic_benchmarks}
\makebox[\textwidth]{%
\begin{tabular}{|c|c|c|c|}
\hline
\textbf{Source English} & ENG-MMLU \cite{mmlu} & EXAMS \cite{exams} & ARC-Challenge \cite{arc} \\ \hline
\textbf{Arabic Translated} & ARB-MMLU \cite{mmlu_arabic} & Arabic\_Exam$^{*}$ \cite{hardalov-etal-2020-exams} & Arabic-ARC-Challenge \cite{AlGhafa-Arabic-LLM-Benchmark-Translated} \\ 
\hline \hline
\textbf{Source English} & ARC-Easy \cite{arc} & BOOLQ \cite{boolq} & COPA \cite{copa} \\ \hline
\textbf{Arabic Translated} & Arabic-ARC-Easy \cite{AlGhafa-Arabic-LLM-Benchmark-Translated} & Arabic-BOOLQ \cite{AlGhafa-Arabic-LLM-Benchmark-Translated} & Arabic-COPA \cite{AlGhafa-Arabic-LLM-Benchmark-Translated} \\
\hline \hline
\textbf{Source English} & HELLASWAG \cite{zellers2019hellaswagmachinereallyfinish} & OPENBOOK-QA \cite{openbookqa} & PIQA \cite{piqa} \\ \hline
\textbf{Arabic Translated} & Arabic-HELLASWAG \cite{AlGhafa-Arabic-LLM-Benchmark-Translated} & Arabic-OPENBOOK-QA \cite{AlGhafa-Arabic-LLM-Benchmark-Translated} & Arabic-PIQA \cite{AlGhafa-Arabic-LLM-Benchmark-Translated} \\
\hline \hline
\textbf{Source English} & RACE \cite{race} & SCIQ \cite{sciq} & TOXIGEN \cite{toxigen} \\ \hline
\textbf{Arabic Translated} & Arabic-RACE \cite{AlGhafa-Arabic-LLM-Benchmark-Translated} & Arabic-SCIQ \cite{AlGhafa-Arabic-LLM-Benchmark-Translated} & Arabic-TOXIGEN \cite{AlGhafa-Arabic-LLM-Benchmark-Translated} \\ 
\hline
\end{tabular}}
\end{table*}

Consider ENG-MMLU \cite{mmlu}, one of the most widely used English benchmarks for evaluating the capabilities of LLMs. It has historically featured on HuggingFace’s Open LLM Leaderboard\cite{open-llm-leaderboard-v2}, serving as a primary dataset for tracking model performance across a range of subjects. Despite its popularity, ENG-MMLU has been shown to contain numerous ground truth errors. Recent efforts like MMLU-Redux \cite{gema2024mmlu} have addressed some of these issues by proposing a hierarchical taxonomy of errors and manually re-annotating a subset of problematic questions. MMLU-Pro\cite{wang2024mmlu}, which has since replaced the ENG-MMLU on the leaderboard, further enhances the benchmark by focusing on more reasoning-intensive questions and expanding the answer choices.

The ARB-MMLU inherits many of the same issues present in the original dataset, including unclear questions and answer choices, incorrect or missing ground truths, and cases with multiple valid answers \cite{gema2024mmlu}. Additionally, the process of machine translation introduces additional challenges such as grammatical inconsistencies, semantic shifts, and cultural misalignment. For instance, the idiom "He is a big shot gun" which is machine-translated as "\<إنه بندقية كبيرة>", which is a literal translation of a shot gun", actually means "\<هو شخص مهم>" in Arabic \cite{ar-translation}, meaning an important person. Such cases highlight the difficulty of translating idiomatic expressions, which are often culturally specific and prone to semantic distortion when subjected to literal translation, especially if the translation overlooks contextual and cultural nuances.

Despite these limitations, there have been no major efforts to systematically revise or improve the ARB-MMLU. To address this gap, we present in Section~\ref{sec:arabic_mmlu} an automatic and systematic approach for refining Enhanced ARB-MMLU datasets, targeting mapping, translation, and content errors, and resulting in an improved, more reliable dataset. Then, in Section \ref{sec:Saudi_Culture}, we expand the Arabic benchmark landscape by introducing the Saudi Culture Dataset, developed in-house to evaluate model alignment with culturally specific knowledge and values relevant to Saudi Arabia and the Gulf region. This addresses a critical gap, as there is currently no benchmark specifically designed for the Saudi context.

\noindent \subsubsection{ARB-MMLU: Diagnostic Challenges and Corrective Approaches} \label{sec:arabic_mmlu}

In this section, we aim to evaluate and improve the quality of ARB-MMLU as defined in Section 4.1. data by analyzing model behavior on a refined version of the ARB-MMLU. This version has been systematically enhanced by correcting ground truth errors, resolving mistranslations, and enforcing consistent evaluation protocols. The goal is to ensure fair and accurate assessment of Arabic NLP tasks using LLMs.

This work focuses on adapting the ARB-MMLU benchmark, which retains the original's size ($\sim$15,000 multiple-choice questions) and was translated using GPT-3.5 Turbo. Efforts were made to ensure dataset accuracy through error correction and consistent evaluation, enabling fair and reliable assessment of Arabic language models.

\begin{figure}[t]
    \centering
        \includegraphics[width=0.95\linewidth]{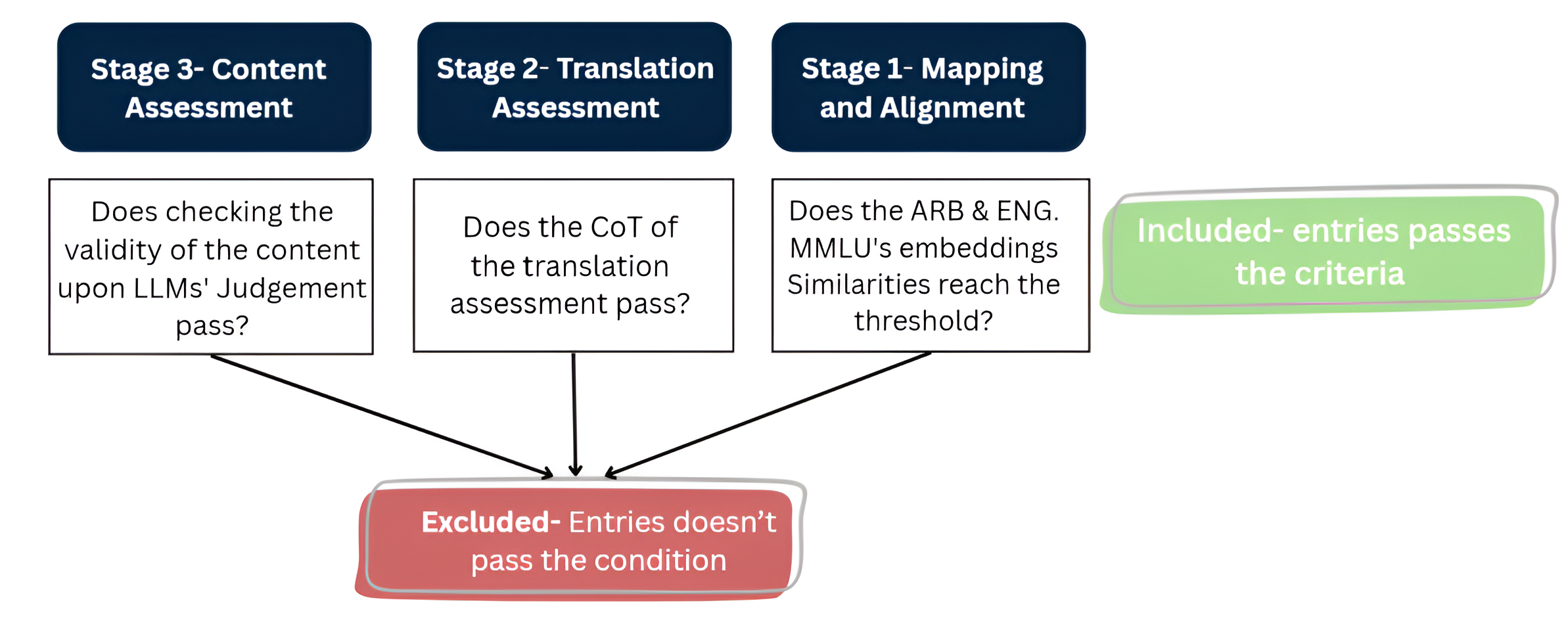}
        \caption{Framework for evaluating ARB-MMLU quality. The process begins with semantic alignment between ARB-MMLU and ENG-MMLU, followed by translation assessment across multiple dimensions, and concludes with chain-of-thought (CoT) prompts to identify and correct content issues.}
        \label{fig:Flowchart_evaluation}

\end{figure}

\begin{itemize}
\item \textbf{Stage 1: Mapping and Alignment.} We semantically align ARB-MMLU items with their ENG-MMLU counterparts to ensure accurate correspondence.
\item \textbf{Stage 2: Translation Assessment.} We evaluate the translation quality of the ARB-MMLU across linguistic, contextual, and semantic dimensions.
\item \textbf{Stage 3: Content Assessment.} Following MMLU-Redux \cite{gema2024mmlu}, we apply CoT prompts to detect labeling issues, producing a refined version of ARB-MMLU.
\end{itemize}

This pipeline (Figure~\ref{fig:Flowchart_evaluation}) provides a principled approach for refining Arabic benchmarks, enabling more reliable and fair evaluation of Arabic LLMs.

\paragraph{Stage 1: Mapping and Alignment}A key challenge is the lack of clear indexing between ARB-MMLU its English counterpart. We work with two sources: (1) the ENG-MMLU, (2) ARB-MMLU, a direct translation of the full English version. 

Our alignment process translates ARB-MMLU entities back into ENG-MMLU, encodes all items with sentence embeddings, and computes cosine similarity to identify best matches. We compared two widely used embedding models: \textit{All-MiniLM-L6-v2} \cite{sentence_transformers_all_minilm_l6_v2}, known for efficient sentence-level similarity, and \textit{BERT} \cite{huggingface_bert_doc}, known for strong contextual embeddings. Mapping accuracy was measured as the percentage of correctly aligned pairs based on Eng-MMLU to ARB-MMLU ground-truth mapped entities.

\noindent \textit{Experimental Findings} Across 10 random-selection topics \ref{tab:average_similarity}, \textit{All-MiniLM-L6-v2} consistently outperformed \textit{BERT}, achieving alignment accuracies ranging from 75.8\% to 100\%, while \textit{BERT} averaged only 39\%. These results confirm \textit{All-MiniLM-L6-v2} as the more effective and efficient model for semantic alignment, as summarized in Table~\ref{tab:average_similarity}. To further improve reliability, we established a cosine similarity threshold of \textbf{0.4686}, below which pairs were consistently misaligned and therefore excluded. We used 83–99 entities from every topic as a robust basis for estimating this threshold and supporting the overall robustness of our alignment pipeline.

\begin{table*}[t]
\centering
\caption{Overall comparison of similarity scores and translation quality across selected ARB-MMLU topics. The results highlight both the degree of semantic alignment between English and Arabic versions (via cosine similarity) and the effectiveness of human translation quality assessments for questions and answers. Together, these tables provide complementary perspectives on dataset reliability and cross-lingual consistency.}
\label{tab:main_comparison}

\begin{subtable}{0.48\textwidth}
    \centering
    \small
    \resizebox{\textwidth}{!}{%
    \begin{tabular}{ccc}
    \toprule
    \textbf{Topic} & \textbf{Mean Cosine Similarity} & \textbf{Number of QA Pairs} \\
    \hline\hline
    Anatomy & 0.805 & 93 \\
    Astronomy & 0.815 & 99 \\
    Business ethics & 0.877 & 97 \\
    Clinical knowledge & 0.892 & 96 \\
    College chemistry & 0.877 & 90 \\
    College Math & 0.864 & 94 \\
    Computer science & 0.837 & 84 \\
    Conceptual physics & 0.845 & 94 \\
    Logical fallacies & 0.734 & 95 \\
    Virology & 0.904 & 95 \\
    \bottomrule
    \end{tabular}%
    }
    \caption{Mean cosine similarity scores across topics, indicating the average semantic closeness between English and Arabic question–answer pairs. Higher scores reflect stronger alignment, with Virology and Clinical Knowledge achieving the strongest similarity.}
    \label{tab:average_similarity}
\end{subtable}
\hfill
\begin{subtable}{0.48\textwidth}
    \centering
    \small
    \resizebox{\textwidth}{!}{%
    \begin{tabular}{p{2.5cm} >{\centering\arraybackslash}p{2cm} >{\centering\arraybackslash}p{2cm} >{\centering\arraybackslash}p{2cm}}
    \toprule
    \textbf{Topic} & \textbf{Questions (\%)} & \textbf{Answers (\%)} & \textbf{Overall (\%)} \\
    \midrule\midrule
    Anatomy & 80.65 & 83.87 & 74.19 \\
    Astronomy & 88.35 & 95.15 & 84.47 \\
    Business Ethics & 87.63 & 89.69 & 84.54 \\
    Clinical Knowledge & 91.67 & 91.67 & 87.50 \\
    College Chemistry & 88.89 & 91.11 & 86.67 \\
    College CS & 85.71 & 92.86 & 84.52 \\
    College Math & 88.30 & 98.94 & 88.30 \\
    Conceptual Physics & 74.47 & 92.55 & 74.47 \\
    Logical Fallacies & 81.05 & 89.47 & 76.84 \\
    Virology & 97.89 & 96.84 & 95.79 \\
    \bottomrule
    \end{tabular}%
    }
    \caption{Translation quality assessment across topics, reporting the percentage of correct translations for questions, answers, and overall items. The consistently high scores suggest strong translation reliability, with Virology and Clinical Knowledge exhibiting near-perfect preservation of meaning across languages.}
    \label{tab:translation_all_topics}
\end{subtable}
\end{table*}

\noindent \textbf{Stage 2: Translation Assessments.} Since ARB-MMLU is a direct translation of ENG-MMLU, including its known errors, it is critical to assess translation quality to determine where meaning may have been distorted or degraded, and how these issues may affect downstream evaluation. We evaluate translation quality along three dimensions: (1) accuracy of the question translation, (2) accuracy of the answer option translations, and (3) overall consistency and correctness. Errors can arise in any of these components, such as ambiguous questions, mistranslated options, or divergences in overall meaning. To perform this assessment, we employ a a prompting method that systematically compares ARB-MMLU with the English entities aligned in Stage 1 (See Appendix). We use Gemini 1.5 Flash \cite{google2025gemini15flash}, which prior work identifies as producing the fewest errors in assessing Arabic translation \cite{al-salman2024assessing}. This allows us to localize translation issues and separate them from alignment or labeling errors.

\noindent \textit{Experimental Findings} We define translation accuracy as the proportion of items judged correct across the three evaluation dimensions (question, answer options, overall consistency). Table~\ref{tab:translation_all_topics} reports translation accuracy across 10 ARB-MMLU topics as judged by Gemini 1.5 Flash under our prompt evaluation. Results show substantial variation across domains. Scientific subjects such as \textbf{Virology (95.79\%)}, \textbf{College Math (88.30\%)}, and \textbf{College Chemistry (86.67\%)} achieve the highest accuracies. Their strong performance likely stems from reliance on numbers, symbols, and formulaic expressions that are structurally consistent and easier to translate. In contrast, domains such as \textbf{Logical Fallacies (76.84\%)} and \textbf{Conceptual Physics (74.47\%)} score lower, reflecting challenges in handling long descriptive text and abstract reasoning, where implicit meaning and complex syntax increase the risk of translation errors. These findings indicate that factual and technical subjects generally translate well, whereas conceptual topics remain more error-prone. This metric therefore provides a useful diagnostic signal for identifying areas in ARB-MMLU requiring improved translation handling or manual review.

\paragraph{Stage 3: Content Assessment}
To support large-scale evaluation of Arabic LLM benchmarks, we propose automated error detection using Chain-of-Thought (CoT) \cite{wei2022chain} prompting. This stage builds on the error taxonomy introduced in MMLU-Redux \cite{gema2024mmlu}, which identifies dataset issues such as \textit{ambiguous questions, wrong ground truth answers, missing correct options, multiple correct answers, and unclear phrasing}. Each item is classified as either 1, for error free questions, or 0, for erroneous sample, for each error type component across all the ARB-MLU using LLMs as judges.

We implement this with GPT-4o and Claude-3.5-sonnet as automated judges. Both models are prompted in Arabic and required to explain their reasoning before making a final classification, following best practices from a recent Arabic benchmark evaluations study \cite{huggingface_blog_3c3h_aragen}. This setup enables scalable and reproducible identification of content issues beyond translation quality.

To further stress-test this framework, we evaluate the ability of LLMs to detect artificially introduced syntactic and content-level errors. We inject controlled errors into the dataset, targeting a subset of ARB-MMLU items already aligned with ENG-MMLU and passed the translation assessment test. For this, we select \textit{Anatomy} and \textit{Clinical Knowledge}, which are considered largely error-free as MMLU-Redux \cite{gema2024mmlu} states. The five error types \textit{Bad Question Clarity, Bad options clarity, No correct answer, Multiple correct answers, and Wrong groundtruth} are randomly applied to 10\% of questions in each topic, with proportional distribution across types. This allows us to assess the sensitivity and consistency of GPT-4o and Claude-3.5-sonnet in detecting benchmark flaws. By applying the CoT-based evaluation template, we measure their ability to identify both superficial distortions and deeper content inconsistencies in the Arabic benchmark.

\begin{figure}[t]
    \centering
    \begin{subfigure}{0.35\linewidth}
        \centering
        \includegraphics[width=0.90\linewidth]{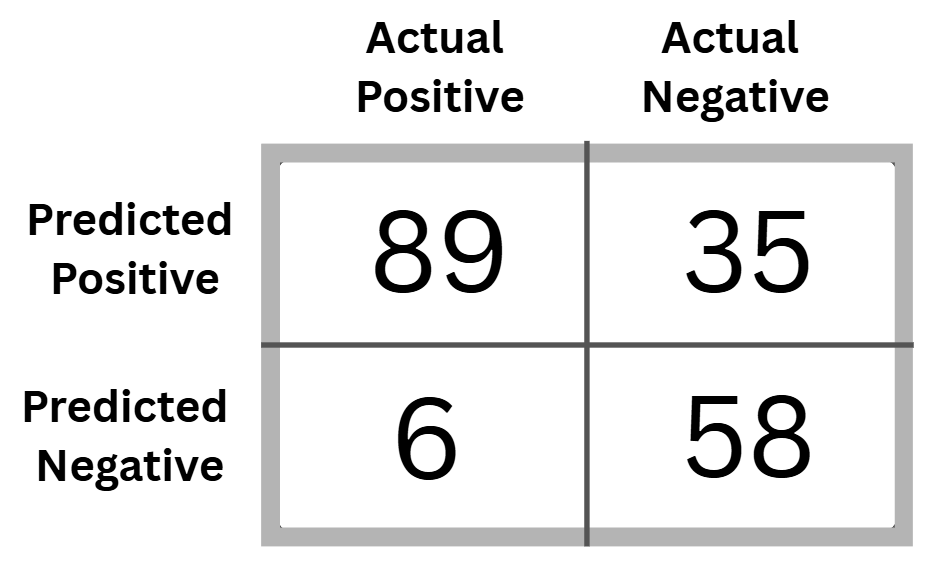}
        % \caption{GPT-4o Error Detection Rate}
        \label{fig:gpt4o-results}
    \end{subfigure}
    \hspace{0.1\linewidth} % Replaced \hfill with a specific space
    \begin{subfigure}{0.35\linewidth}
        \centering
        \includegraphics[width=0.90\linewidth]{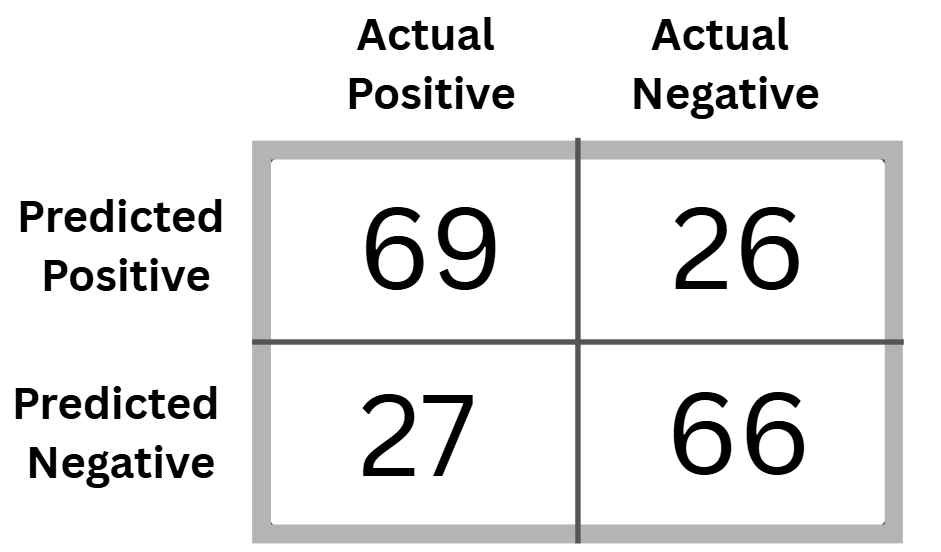}
        % \caption{Claude Sonnet Error Detection Rate}
        \label{fig:sonnet-results}
    \end{subfigure}
    \caption{Comparison of error detection performance by GPT-4o (left) and Claude-3.5 Sonnet (right) on the ARB-MMLU subset with artificially injected errors.}
    \label{fig:error-detection-comparison}
\end{figure}

\noindent \textit{Injected Error–Detection Performance} To evaluate the reliability of LLMs in detecting content-level errors, we tested GPT-4o and Claude-3.5 Sonnet on the subset of ARB-MMLU with systematically injected errors. Table~\ref{tab:combined-stats} reports precision, recall, false negative rate, and F1 scores for both positive (error) and negative (clean) predictions.
The results show that GPT-4o achieves higher recall and stronger overall accuracy, reflecting its ability to capture a larger proportion of injected errors. Claude-3.5 Sonnet, while more conservative, reduces false positives at the cost of missing more errors. This trade-off suggests different deployment contexts: GPT-4o is preferable in recall-driven scenarios where identifying as many errors as possible is critical, while Claude-3.5 Sonnet is more suitable when minimizing incorrect error flags is prioritized. These findings challenge the assumption that Claude-3.5 Sonnet is universally the stronger evaluator \cite{huggingface_blog_3c3h_aragen}, underscoring the importance of balancing recall and precision in benchmark evaluation.
\begin{table}[t]
\centering
\resizebox{\textwidth}{!}{
\begin{tabular}{l | c c c c c !{\color{gray}\vrule} c c c c c }
\specialrule{.1em}{0em}{0em} 
\multirow{2}{*}{\textbf{Model Name}} 
& \multicolumn{5}{c !{\color{gray}\vrule}}{\textbf{GPT-4o}} 
& \multicolumn{5}{c}{\textbf{Claude-3.5 Sonnet }} \\
\cline{2-11}
& \textbf{Precision} & \textbf{1-Precision} & \textbf{Recall} & \textbf{FNR} & \textbf{F1} 
& \textbf{Precision} & \textbf{1-Precision} & \textbf{Recall} & \textbf{FNR} & \textbf{F1} \\
\specialrule{.1em}{0em}{0em} 
Predict (Positive) & 0.9368 & 0.0632 & 0.7177 & 0.2823 & 0.8128 
& 0.7188 & 0.2813 & 0.7263 & 0.2737 & 0.7225 \\
Predict (Negative) & 0.6237 & 0.3763 & 0.9063 & 0.0938 & 0.7389 
& 0.7174 & 0.2826 & 0.7097 & 0.2903 & 0.7135 \\
\specialrule{.1em}{0em}{0em} 
\end{tabular}
}
\caption{The performance of models in detecting deliberately injected content errors in the ARB-MMLU dataset is evaluated. The results include metrics like precision, recall, false negative rate, and F1 score for both error (positive) and clean (negative) predictions, reflecting model's reliability in identifying these errors.
}
\label{tab:combined-stats}
\end{table}

\begin{table}[t]
\centering
% \small
\resizebox{\textwidth}{!}{%
\begin{tabular}{lccccc}
\textbf{Topic} & \textbf{Question Presentation (\%)} & \textbf{MC Options Presentation (\%)} & \textbf{Answer Evaluation (\%)} & \textbf{Ground Truth Answer (\%)} & \textbf{Overall Classification (\%)} \\
\hline\hline
Astronomy & 88.7 & 85.4 & 87.1 & 86.3 & 89.0 \\
Business ethics & 83.6 & 81.7 & 84.2 & 82.9 & 83.1 \\
College chemistry & 80.0 & 83.5 & 79.9 & 82.1 & 81.7 \\
College Math & 69.1 & 72.3 & 70.8 & 71.5 & 68.9 \\
Computer science & 77.90 & 75.6 & 78.30 & 80.2 & 76.8 \\
Logical fallacies & 74.4 & 69.8 & 72.5 & 75.2 & 70.1 \\
Professional law & 65.3 & 70.0 & 68.7 & 69.4 & 67.2 \\
Virology & 85.2 & 78.9 & 90.1 & 88.5 & 82.3 \\
\hline
\end{tabular}%
}
\caption{Translation evaluation accuracy across multiple-choice question components for 8 validation topics. Scores reflect accuracy of (i) question translation, (ii) multiple-choice options translation, (iii) predicted answers, (iv) ground-truth answer alignment, and (v) overall classification. Results highlight that while question translations remain relatively reliable, translation fidelity for short answer options is more error-prone.}
\label{tab:random_evaluation}
\end{table}

\begin{figure}[t]
    \centering
    \includegraphics[width=1\linewidth]{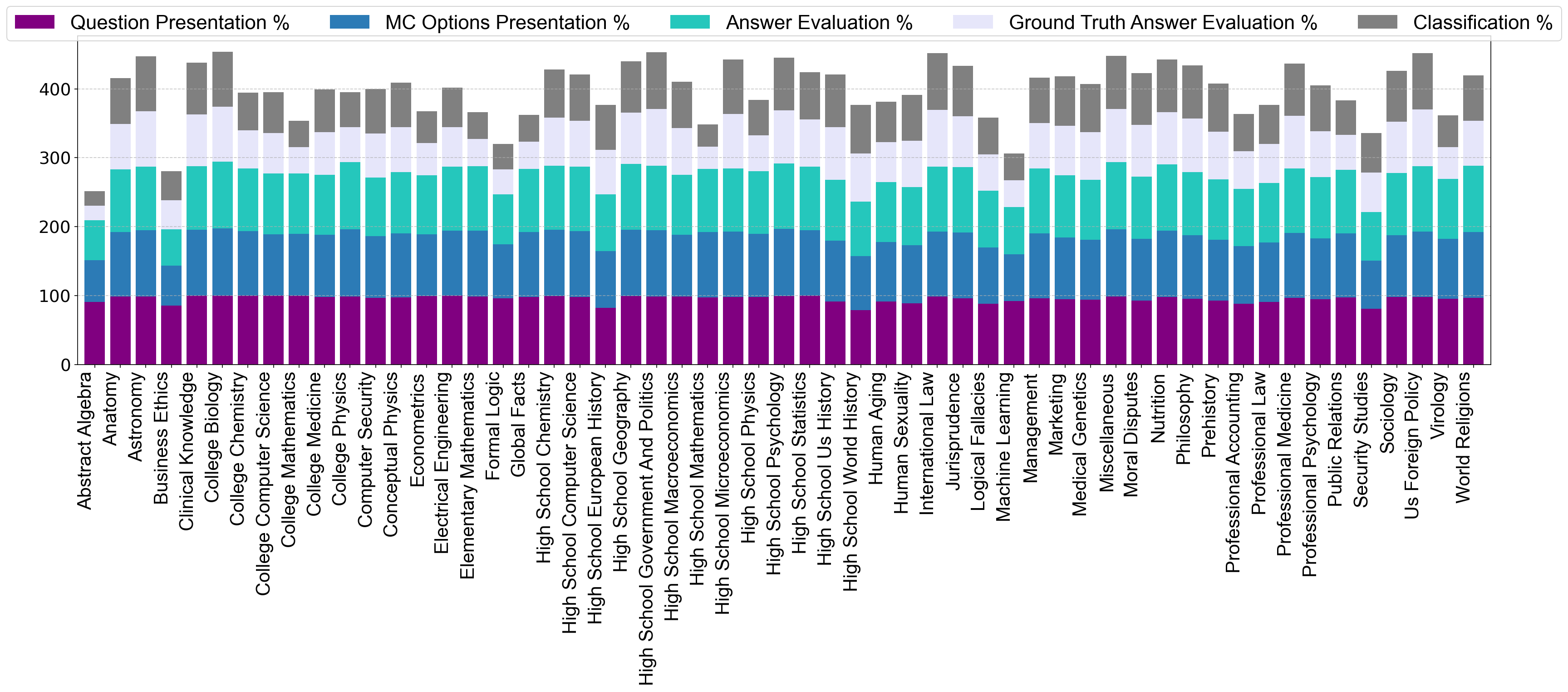}
    \caption{
    Distribution of flagged error types across ARB-MMLU topics using GPT-4. Each stacked bar shows the proportion of items per topic classified as Question Presentation, MC Options Presentation, Answer Evaluation, Ground Truth Answer Evaluation, or Classification. The Figure shows variation in error prevalence across different subject areas.}
    \label{fig:all_assessed}
\end{figure}

\noindent \textit{Generalization of Correction Results.} Our analysis reveals that content quality remains relatively high for the \textit{question presentation} component across most topics (Fig~\ref{fig:all_assessed}). However, accuracy declines more noticeably when evaluating the \textit{answer options}. This discrepancy likely arises because longer, context-rich question texts provide models with sufficient semantic cues to ground their reasoning, while short and often ambiguous answer choices lack such context. As a result, models struggle to capture the intended meaning of these compact options, making them harder to interpret and evaluate faithfully.

Among all topics, \textbf{Astronomy} and \textbf{Logical Fallacies} achieved stronger classification accuracy, suggesting either that translations in these domains preserved meaning more effectively or that their reasoning requirements were easier for LLMs to evaluate in Arabic. In contrast, the \textbf{Mathematics} domain exhibited the lowest performance, reflecting the sensitivity of math problems to precise numerical values that are easily distorted during translation. Similarly, \textbf{Mathematics} and \textbf{Chemistry} showed the largest drops in \textit{Ground Truth Answer Evaluation}, indicating that translation mismatches or misinterpretations of specialized notation (e.g., equations, formulas, units) played a key role in producing misleading evaluations.

These findings underscore the need for robust handling of domain-specific technical content in translation, particularly for symbolic, mathematical, or formulaic expressions. Enhancing translation fidelity in such domains is crucial for ensuring the reliability and fairness of Arabic-language benchmarks. Importantly, we generalized this process across all 114 ARB-MMLU topics, covering both validation and test sets, which enabled us to systematically record and analyze every error type flagged at each stage of evaluation (Table~\ref{tab:random_evaluation}).

In summary, our pipeline strengthens the reliability of Arabic benchmark datasets by addressing issues of alignment, translation quality, and content integrity. The proposed three-stage approach: (1) semantic matching with MMLU-Redux aligned English data, (2) translation quality assessment using Gemini 1.5 Flash, and (3) content error detection with GPT-4o and Claude-3.5 Sonnet, which provides a scalable framework for dataset refinement. Beyond Arabic, our methodology can be  applied towards fairer and more accurate multilingual evaluation.

Bringing these three stages together, we construct a refined dataset that excludes problematic entries and achieves higher overall quality. This detoxification process improved the benchmark ARB-MMLU by addressing both translation issues and content-level errors present in the original dataset. Figure~\ref{fig:all_assessed} summarizes the statistics of this new dataset, highlighting reductions in identified errors and improvements in alignment fidelity. Concretely, the ARB-MMLU-Test set was reduced from 14,042 samples to 6,804, and the ARB-MMLU-Dev set from 285 samples to 127, reflecting a substantial refinement in dataset reliability and usability across categories.

In the next section, we introduce the Saudi Culture Dataset, created in-house to evaluate model alignment with cultural knowledge and values specific to Saudi Arabia and the Gulf region. This dataset complements existing Arabic benchmarks by addressing a previously unrepresented cultural context in evaluation.

\begin{table}[t]
\centering
\resizebox{\columnwidth}{!}{%
\begin{tabular}{ccccccccc}
\hline
\multicolumn{1}{c|}{\textbf{Category}} &
  \multicolumn{1}{c|}{\textbf{History}} &
  \multicolumn{1}{c|}{\textbf{Tradition\&Customs}} &
  \multicolumn{1}{c|}{\textbf{Art\&Architecture}} &
  \multicolumn{1}{c|}{\textbf{Cuisine}} &
  \multicolumn{1}{c|}{\textbf{Music\&Dance}} &
  \multicolumn{1}{c|}{\textbf{Language}} &
  \multicolumn{1}{c|}{\textbf{Festivals}} &
  \textbf{Sports} \\ \hline
\multicolumn{1}{c|}{\multirow{4}{*}{\textbf{Subcategories}}} &
  \multicolumn{1}{c|}{Pre-Islamic History} &
  \multicolumn{1}{c|}{Bedouin Heritage} &
  \multicolumn{1}{c|}{Traditional Art} &
  \multicolumn{1}{c|}{Traditional Dishes} &
  \multicolumn{1}{c|}{Traditional Music} &
  \multicolumn{1}{c|}{Arabic Language} &
  \multicolumn{1}{c|}{Eid alFitr\&alAdha} &
  Football \\
\multicolumn{1}{c|}{} &
  \multicolumn{1}{c|}{Islamic History} &
  \multicolumn{1}{c|}{Family\&Social Structure} &
  \multicolumn{1}{c|}{Modern Art} &
  \multicolumn{1}{c|}{Drinks} &
  \multicolumn{1}{c|}{Dance} &
  \multicolumn{1}{c|}{Poetry} &
  \multicolumn{1}{c|}{Saudi celebration Days} &
  Camel Racing \\
\multicolumn{1}{c|}{} &
  \multicolumn{1}{c|}{The Kingdom of Saudi Arabia} &
  \multicolumn{1}{c|}{Traditional Clothing} &
  \multicolumn{1}{c|}{Architecture} &
  \multicolumn{1}{c|}{Sweets} &
  \multicolumn{1}{c|}{} &
  \multicolumn{1}{c|}{Accents} &
  \multicolumn{1}{c|}{Religious Pilgrimages} &
  Falconry \\
\multicolumn{1}{c|}{} &
  \multicolumn{1}{c|}{Gulf States’ History} &
  \multicolumn{1}{c|}{Social Etiquette} &
  \multicolumn{1}{c|}{} &
  \multicolumn{1}{c|}{} &
  \multicolumn{1}{c|}{} &
  \multicolumn{1}{c|}{Proverbs and Sayings} &
  \multicolumn{1}{c|}{} &
  Horse Racing \\ \hline
\end{tabular}%
}
\caption{Our proposed Saudi Culture Dataset is structured into eight main categories spanning domains from History, Traditions to Festivals and Sports, each with multiple subcategories (e.g., Pre-Islamic History, Bedouin Heritage, Traditional Art, Arabic Poetry, Camel Racing).The dataset includes 350 multi-turn questions reflecting diverse aspects of Saudi cultural}
\label{tab:saudi-table-size}
\end{table}

\subsubsection{New Dataset: Saudi Culture Dataset} \label{sec:Saudi_Culture}

While several datasets have been developed to evaluate the cultural competence of LLMs, many focus on global or generalized cultural settings. Benchmarks such as GeoMLAMA \cite{geomlama}, CultureAtlas \cite{CultureAtlas}, and StereoKG \cite{stereokg} attempt to capture culturally relevant knowledge but primarily rely on English-language data or reflect Western-centric norms.

Within the Arabic NLP landscape, a number of culturally focused benchmarks have emerged. Datasets such as ArabCulture \cite{ArabCulture}, AraDiCE \cite{aradice}, and CIDAR \cite{cidar} have begun addressing cultural and dialectal nuances in Arabic-speaking populations. Yet, these efforts often treat the Arab world as culturally uniform, emphasizing Modern Standard Arabic (MSA) while overlooking region-specific practices. In particular, the Gulf region, and Saudi Arabia specifically, remains critically underrepresented.

To fill this gap, we introduce the Saudi Culture Dataset, a culturally grounded benchmark specifically designed to evaluate Arabic LLMs on their understanding of Saudi and Gulf cultural contexts. This dataset enriches evaluation by incorporating culturally specific questions that reflect local customs, values, and societal behaviors. Table~\ref{tab:saudi-table-size} outlines the dataset’s main categories and subcategories.

\textbf{Saudi Culture Dataset Construction.} We build the dataset from three sources: a translated and culturally adapted subset of MT-Bench \cite{mt-bench}, a filtered portion of Pico-Saudi LLMs Benchmark \cite{pico-Q}, and a large collection of newly created Saudi cultural questions that we designed in-house. The first two sources provide smaller but useful baselines (45 out of 80 and 35 out of 55 as a total questions, respectively), ensuring coverage of general conversational and Saudi-specific evaluation. However, the core of the dataset 270 questions, about 77\% of the overall 350 comes from our manually generated Saudi cultural questions.\par

To create these, we first defined a taxonomy of main and subcategories of Saudi cultural knowledge spanning traditions, social norms, and everyday practices (see Table \ref{tab:saudi-dataset} for the complete list of categories). For each subcategory, we authored two multi-turn questions across diverse task types (writing, roleplay, extraction, reasoning, humanities). Question generation was performed using ChatGPT as a drafting tool, but every item was carefully curated, adapted, and refined by our team to ensure cultural authenticity and linguistic naturalness. This process produced 270 multi-turn questions, representing 77\% of the dataset and forming its distinctive contribution.

{
\scriptsize
\begin{longtable}{>{\centering\arraybackslash}m{2cm}|
                  >{\centering\arraybackslash}m{2cm}|
                  >{\centering\arraybackslash}m{1.6cm}|
                  >{\centering\arraybackslash}m{7.5cm}}

\hline
\textbf{Category} & \textbf{Sub-Category} & \textbf{Question Type} & \multicolumn{1}{c}{\textbf{Example}} \\ \hline
\endfirsthead

\hline
Category & Sub-Category & Question Type & \multicolumn{1}{c|}{Example} \\ \hline
\endhead

\hline \multicolumn{4}{|r|}{{Continued on next page}} \\ \hline
\endfoot

\endlastfoot

\textbf{History} & Pre-Islamic History & Writing & \begin{arabtext}
اكتب مقالًا عن معركة بدر وأثرها في تاريخ العرب قبل الإسلام. قارنها بمعركة أخرى. 
أعد كتابة المقال مع التركيز على دور التحالفات القبلية في المعركة.
\end{arabtext} \\ 

\textbf{Tradition\&Customs} & Bedouin Heritage & Humanities & \begin{arabtext}
كيف ساهمت الحياة البدوية في تشكيل القيم الاجتماعية مثل الكرم والشجاعة في المجتمع السعودي؟
تخيل أنك تسافر إلى مجتمع بدوي معاصر، كيف ستشهد تطبيق هذه القيم في الحياة اليومية؟
\end{arabtext} \\ 

\textbf{Art \& Architecture} & Traditional Art & Reasoning & \begin{arabtext}
إذا كانت اللوحة تحتوي على عشرة أشكال هندسية، وقام الفنان بإضافة نصف العدد مرة أخرى، كم عدد الأشكال الآن؟
المرجع: سيكون هناك 15 شكلًا.
\end{arabtext} \\ 

\textbf{Cuisine} & Traditional Dishes & Roleplay & \begin{arabtext}
أنت الآن مالك لمقهى سعودي تقليدي في جدة، ويطلب منك الزبون مشروب السوبيا الذي يتم تحضيره بطريقة خاصة. كيف ستشرح له طريقة تحضيره؟
\end{arabtext} \\ 

\textbf{Music\&Dance} & Dance & Writing & \begin{arabtext}
كيف ترى العلاقة بين الموسيقى والرقصات الشعبية السعودية؟ هل تعتقد أن الإيقاع يعكس جزءًا من الهوية الثقافية للمجتمع السعودي؟
\end{arabtext} \\ 

\textbf{Language} & Poetry & Extraction & \begin{arabtext}
اقرأ أبيات الشعر التالية للمتنبي وحدد الغرض الشعري الأبرز في الأبيات:
إذا غامرت في شرف مروم ** فلا تقنع بما دون النجوم
فطعم الموت في أمر صغير ** كطعم الموت في أمر عظيم
المرجع: الشجاعة والطموح
\end{arabtext} \\ 

\textbf{Festivals} & Saudi Celebration Days & Roleplay & \begin{arabtext}
تخيل أنك شخصية تاريخية عاشت في فترة تأسيس المملكة العربية السعودية. كيف ستروي تجربتك في تلك الحقبة وكيف ساهمت في تأسيس المملكة؟
\end{arabtext} \\ 

\textbf{Sports} & Camel Racing & Humanities & \begin{arabtext}
قارن بين تنظيم سباقات الهجن في السعودية والبلدان العربية الأخرى. كيف يؤثر هذا التنظيم على تطوير السياحة الرياضية في البلد؟
\end{arabtext}\\ \hline
\caption{Representative samples from our Saudi Culture Dataset illustrate how each category and subcategory is paired with diverse task types (Writing, Roleplay, Reasoning, Humanities, Extraction). The examples showcase Arabic multi-turn questions that evaluate both factual knowledge (e.g., history, cuisine, sports) and interpretive reasoning (e.g., social customs, poetry), highlighting the dataset’s role in testing LLMs’ cultural competence in Saudi and Gulf contexts.} 
\label{tab:saudi-dataset} 
\end{longtable}
}

\subsection{Leaderboard and Full New Evaluation} \label{sec:leaderboard}

This section introduces our new leaderboard for evaluating Arabic LLMs. The goal is to provide a comprehensive comparison of Arabic models across knowledge, reasoning, and cultural understanding, using both established and newly created benchmark. In particular, we highlight two contributions: (i) our proposed Enhanced-ARB-MMLU benchmark, a cleaned variant of the original ARB-MMLU, and (ii) an answer-shuffling protocol that diagnoses sensitivity to superficial formatting.

\begin{table}[t]
\centering
\fontsize{22}{26}\selectfont
\resizebox{\textwidth}{!}{
\begin{tabular}{p{9cm} c c c c c c c c c !{\color{gray}\vrule} c c c  c c c}
\specialrule{.1em}{0em}{0em} 
\multirow{3}{*}{\textbf{Model Name (Size)}} 
& \multicolumn{3}{c}{\textbf{Alghafa}} 
& \multicolumn{3}{c}{\textbf{ACVA}} 
& \multicolumn{3}{c !{\color{gray}\vrule}}{\textbf{Arabic\_Exams}} 
& \multicolumn{3}{c}{\textbf{ARB-MMLU}} 
& \multicolumn{3}{c}{\textbf{Enhanced-ARB-MMLU (Ours)}}\\
\cline{2-16}
& \textbf{Lighteval} & \multicolumn{2}{c}{\textbf{Shuffle}} 
& \textbf{Lighteval} & \multicolumn{2}{c}{\textbf{Shuffle}} 
& \textbf{Lighteval} & \multicolumn{2}{c !{\color{gray}\vrule}}{\textbf{Shuffle}} 
& \textbf{Lighteval} & \multicolumn{2}{c}{\textbf{Shuffle}} 
& \textbf{Lighteval} & \multicolumn{2}{c}{\textbf{Shuffle}} \\
&  & Avg & ± std &  & Avg & ± std &  & Avg & ± std &  & Avg & ± std &  & Avg & ± std \\
\specialrule{.1em}{0em}{0em} 
AceGPT (7B) & 52.55 & 53.06 & 0.47 & 71.14 & 68.77 & 2.12 & 26.07 & 26.29 & 0.19 & 26.95 & 26.33 & 0.55 & 27.19 &\textbf{ 26.93} & 0.23 \\
Mistral-v0.3 (7B) & 58.32 & 58.49 & 0.15  & 62.99 & 57.39 & 5.01 & 31.84 & 30.76 & 0.97 & 34.24 & 33.69 & 0.49 & 39.34 & \textbf{37.91 }& 1.28 \\
Gemma (7B) & 69.14 & 69.17 & 0.03 & 63.54 & 64.47 & 0.84 & 45.07 & 45.50 & 0.38 & 48.28 & 47.58 & 0.63  & 58.10 & \textbf{57.86} & 0.21 \\
Qwen-2.5 (7B)  & 72.30 & 72.54 & 0.22  & 82.29 & 82.67 & 0.34 & 48.60 & 48.31 & 0.26 & 53.00 & 52.60 & 0.36 & 68.41 & \textbf{66.82} & 1.42 \\
Llama-3 (8B) & 64.88 & 64.88 & 0.00  & 77.42 & 77.41 & 0.01 & 40.04 & 40.04 & 0.00 & 44.99 & 44.28 & 0.64 & 53.21 & \textbf{52.54} & 0.59 \\
Jais (13B) & 50.08 & 49.98 & 0.09 & 59.77 & 61.51 & 1.56 & 26.63 & 29.09 & 2.20 & 28.51 &\textbf{28.67} & 0.14  & 23.10 & 25.04 & 1.73\\
Llama-2 (13B) & 52.40 & 53.11 & 0.64 & 67.66 & 66.88 & 0.70 & 26.07 & 27.16 & 0.97 & 28.00 & 28.18 & 0.16 & 31.23 & \textbf{29.65} & 1.41\\
\hline
\end{tabular}
}
\caption{The Arabic LLM leaderboard comparing state-of-the-art pre-trained LLMs across multiple Arabic benchmarks. The evaluation covers five datasets: Alghafa, ACVA, Arabic Exams, ARB-MMLU, and Enhanced-ARB-MMLU. The table reports model performance across these benchmarks, reflecting each model's capability in Arabic.}
\label{tab:leaderboard_pretrained}
\end{table}

\begin{table}[t]
\centering
\fontsize{22}{26}\selectfont
\resizebox{\textwidth}{!}{
\begin{tabular}{p{9cm} c c c c c c c c c !{\color{gray}\vrule}  c c c c c c}
\specialrule{.1em}{0em}{0em} 
\multirow{3}{*}{\textbf{Model Name (Size)}} 
& \multicolumn{3}{c}{\textbf{Alghafa}} 
& \multicolumn{3}{c}{\textbf{ACVA}} 
& \multicolumn{3}{c !{\color{gray}\vrule}}{\textbf{Arabic\_Exams}} 
& \multicolumn{3}{c}{\textbf{ARB-MMLU}} 
& \multicolumn{3}{c}{\textbf{Enhanced-ARB-MMLU (Ours)}}\\
\cline{2-16}
& \textbf{Lighteval} & \multicolumn{2}{c}{\textbf{Shuffle}} 
& \textbf{Lighteval} & \multicolumn{2}{c}{\textbf{Shuffle}} 
& \textbf{Lighteval} & \multicolumn{2}{c !{\color{gray}\vrule}}{\textbf{Shuffle}} 
& \textbf{Lighteval} & \multicolumn{2}{c}{\textbf{Shuffle}} 
& \textbf{Lighteval} & \multicolumn{2}{c}{\textbf{Shuffle}} \\
&  & Avg & ± std &  & Avg & ± std &  & Avg & ± std &  & Avg & ± std &  & Avg & ± std \\
\specialrule{.1em}{0em}{0em} 
AceGPT-chat (7B)  & 52.39 & 52.79 & 0.36  & 75.63 & 72.36 & 2.92 & 33.15 & 35.10 & 1.74 & 34.35 & 33.43 & 0.82 & 40.23 & \textbf{39.84 }& 0.35 \\
Mistral-v0.3-Instruct (7B) & 63.63 & 63.62 & 0.01  & 74.17 & 72.45 & 1.53 & 33.71 & 33.20 & 0.45 & 34.73 & 34.51 & 0.20 & 39.42 & \textbf{38.69} & 0.65\\
Gemma-it (7B) & 59.01 & 59.04 & 0.03  & 65.58 & 61.30 & 3.82 & 29.80 & 31.68 & 1.68  & 35.50 & 35.20 & 0.27 & 40.79 & \textbf{40.53} & 0.23\\
Qwen-2.5-Instruct (7B)  & 74.31 & 74.64 & 0.30 & 79.73 & 80.18 & 0.41 & 50.84 & 50.84 & 0.00 & 54.76 & 54.30 & 0.41  & 68.71 & \textbf{67.98} & 0.65 \\
ALLaM-Instruct* (7B) & 69.51 & 69.72 & 0.19  & 77.64 & 78.28 & 0.58 & \textbf{54.00} & \textbf{53.57} & \textbf{0.38} & 52.33 & 52.07 & 0.23 & 66.56 &\textbf{ 66.53} & 0.03\\
Llama-3-Instruct (8B)  & 69.63 & 69.92 & 0.26  & 79.55 & 79.71 & 0.14 & 43.20 & 43.13 & 0.06 & 44.11 & 43.61 & 0.45  & 53.90 & \textbf{53.72 }& 0.16\\
Aya-23* (8B) & 67.64 & 67.65 & 0.01 & 77.47 & 76.43 & 0.93 & 41.34 & 41.70 & 0.32 & 41.50 & 41.33 & 0.15  & 48.22 & \textbf{48.69} & 0.42\\
Yi-1.5-Chat* (9B) & 61.21 & 61.27 & 0.05 & 70.52 & 70.83 & 0.27 & 29.80 & 29.80 & 0.00 & 34.95 & 34.64 & 0.27  & 39.06 & \textbf{39.17} & 0.10 \\
Jais–chat (13B) & 66.13 & 66.17 & 0.04 & 75.24 & 75.36 & 0.10  & 43.58 & 43.50 & 0.07 & 39.96 & 39.93 & 0.03 & 48.77 &\textbf{ 48.93} & 0.14 \\
Llama-2-Instruct (13B) & 48.89 & 48.30 & 0.53 & 67.14 & 65.96 & 1.06 & 27.75 & 27.24 & 0.45  & 28.73 & 28.25 & 0.43 & 30.02 & \textbf{29.27} & 0.67\\
\hline
\end{tabular}
}
\caption{The Arabic LLM leaderboard comparing state-of-the-art fine-tuned LLMs across multiple Arabic benchmarks. The evaluation covers five datasets: Alghafa, ACVA, Arabic Exams, ARB-MMLU, and Enhanced-ARB-MMLU. The table reports model performance across these benchmarks, reflecting each model's capability in Arabic. Models marked with an asterisk (*) have fine-tuned versions only, with no pre-trained versions.}
\label{tab:leaderboard_finetuned}
\end{table}

\begin{table}[t]
    \centering
    \begin{subtable}{0.80\textwidth}
        \centering
        \resizebox{\textwidth}{!}{%
            \begin{tabular}{cccc}
                \toprule
                \textbf{Dataset Name} & \textbf{Original Language} & \textbf{Size} & \textbf{Domain} \\
                \midrule
                Alghafa & English & 22,977 & General Knowledge \\
                Arabic\_Exams & English & 562 & Academic Exams \\
                ACVA & Arabic & 8,370 &  Arabic Culture \\
                ARB-MMLU & Translated Arabic & 14,327 & Education and Knowledge \\
                Enhanced-ARB-MMLU (ours) & Translated Arabic & 6,931 & Education and Knowledge \\
                \bottomrule
            \end{tabular}%
        }
    \end{subtable}
     \caption{Overview of the benchmark datasets used in our evaluation, including their original languages, sizes, and main domains.}
    \label{tab:Benchmarks_1}
\end{table}

\textbf{Datasets.} We evaluated a broad set of Arabic-supporting models, integrating a new benchmark, Enhanced-ARB-MMLU, together with four public datasets: Alghafa \cite{almazrouei-etal-2023-alghafa}, ACVA \cite{ACVA}, Arabic\_Exams \cite{Arabic_EXAMS}, and original ARB-MMLU \cite{mmlu_arabic}. Table~\ref{tab:Benchmarks_1} summarizes all benchmarks used in the evaluation.

\textbf{Evaluation Setup.} Models range from 7B to 13B parameters and include both pre-trained and fine-tuned variants. Representative families include AceGPT \cite{acegpt}, Mistral \cite{mistralai_family}, Gemma \cite{gemma}, Qwen-2.5 \cite{qwen2.5}, ALLaM \cite{allam_paper}, Llama-2 \cite{llama2}, Llama-3 \cite{llama3}, Aya \cite{aya}, Yi \cite{Yi}, and Jais \cite{jais}, ensuring diversity across training paradigms and scales.  

We adopt the LightEval framework \cite{lighteval} in a 5-shot setting, reporting accuracy separately for each benchmark dataset. Models select answers by computing normalized log-likelihood across candidate completions and choosing the highest-scoring option. Most benchmarks use multiple-choice formats (A–D), except ACVA (True/False) and Alghafa (full-string answers). For robustness, we introduce a novel \emph{answer-shuffling} protocol, based on the observation that models may exploit positional biases in candidate options rather than capturing the underlying task semantics. The protocol keeps labels fixed, running one evaluation without shuffling and two more with candidate choices randomly permuted. Results are reported as mean accuracy with standard deviation. This procedure provides additional diagnostic insight into both model robustness and benchmark reliability.

\textbf{Results.} Tables \ref{tab:leaderboard_pretrained} and \ref{tab:leaderboard_finetuned} present the performance of pre-trained and fine-tuned Arabic LLMs across five benchmarks. Among pre-trained models, Qwen-2.5 (7B) achieved the highest accuracy overall, reaching 82.29\% on ACVA and 72.30\% on Alghafa. Gemma (7B) was competitive on Alghafa (69.14\%) but lagged behind on other tasks. Notably, LLaMA-3 (8B) performed strongly on ACVA (77.42\%) while maintaining balanced results across benchmarks. In contrast, larger models such as Jais (13B) and LLaMA-2 (13B) trailed behind, showing that scale alone does not guarantee stronger Arabic performance.
Fine-tuning consistently improved performance. Qwen-2.5-Instruct (7B) achieved the highest accuracy on most benchmarks, including 74.31\% on Alghafa, 79.73\% on ACVA, and 68.71\% on Enhanced-ARB-MMLU. On Arabic\_Exams, ALLaM-Instruct (7B) achieved the highest score of 54.00\%, exceeding Qwen-2.5-Instruct (7B) by more than three points.  LLaMA-3-Instruct (8B) also made large gains over its base model, reaching 79.55\% on ACVA and competitive scores elsewhere, though it still trailed Qwen-2.5-Instruct (7B) overall. Models like Aya-23 (8B) and Yi-1.5-Chat (9B) were moderately strong, while Jais-chat (13B) improved substantially over its base model but did not match smaller fine-tuned models.

A consistent finding is that models achieve significantly higher accuracy on our Enhanced-ARB-MMLU compared to the original ARB-MMLU. This validates our cleaning pipeline, which pruned erroneous and mistranslated samples, and indicates that Arabic LLMs are more capable than previously suggested by noisy benchmark.

The shuffle setup led to noticeable shifts in accuracy. For example, Mistral-v0.3 (7B) dropped by more than 5 points on ACVA, while Qwen-2.5-Instruct (7B) shifted by almost 1 point on Enhanced-ARB-MMLU. In contrast, some models such as Gemma (7B) and Aya-23 (8B) were highly stable across shuffled and non-shuffled settings. These results highlight that Arabic LLMs remain sensitive to option order, underscoring the importance of robustness evaluation.

\begin{tcolorbox}[colback=black!5!white,colframe=black!75!black,title=Conclusion]
We introduced a dedicated leaderboard for assessing Arabic and multilingual models across diverse tasks. Our contributions are twofold: (i) Enhanced-ARB-MMLU, a cleaned, adapted benchmark that provides more reliable evaluation than existing translated datasets, and (ii) a novel answer-shuffling protocol for diagnosing model robustness and benchmark stability. Empirical results show that Enhanced-ARB-MMLU reveals stronger performance than previously reported, highlighting the importance of benchmark quality in evaluating low-resource languages.
\end{tcolorbox}

\bibliographystyle{iclr2024_conference}
\bibliography{refs}

\newpage
\appendix

\label{appendix}

\section{Comparison of text extractors}

This appendix shows an example of the extracted text of each extractor, as discussed in Section 2.2.2 \\
The examples are randomly selected and show Resiliparse as the longest extraction. However, with manual inspection of more examples, you can find that the WET extraction is generally longer and noisier.

  \begin{figure}[h]
    \centering
    \includegraphics[width=0.85\textwidth]{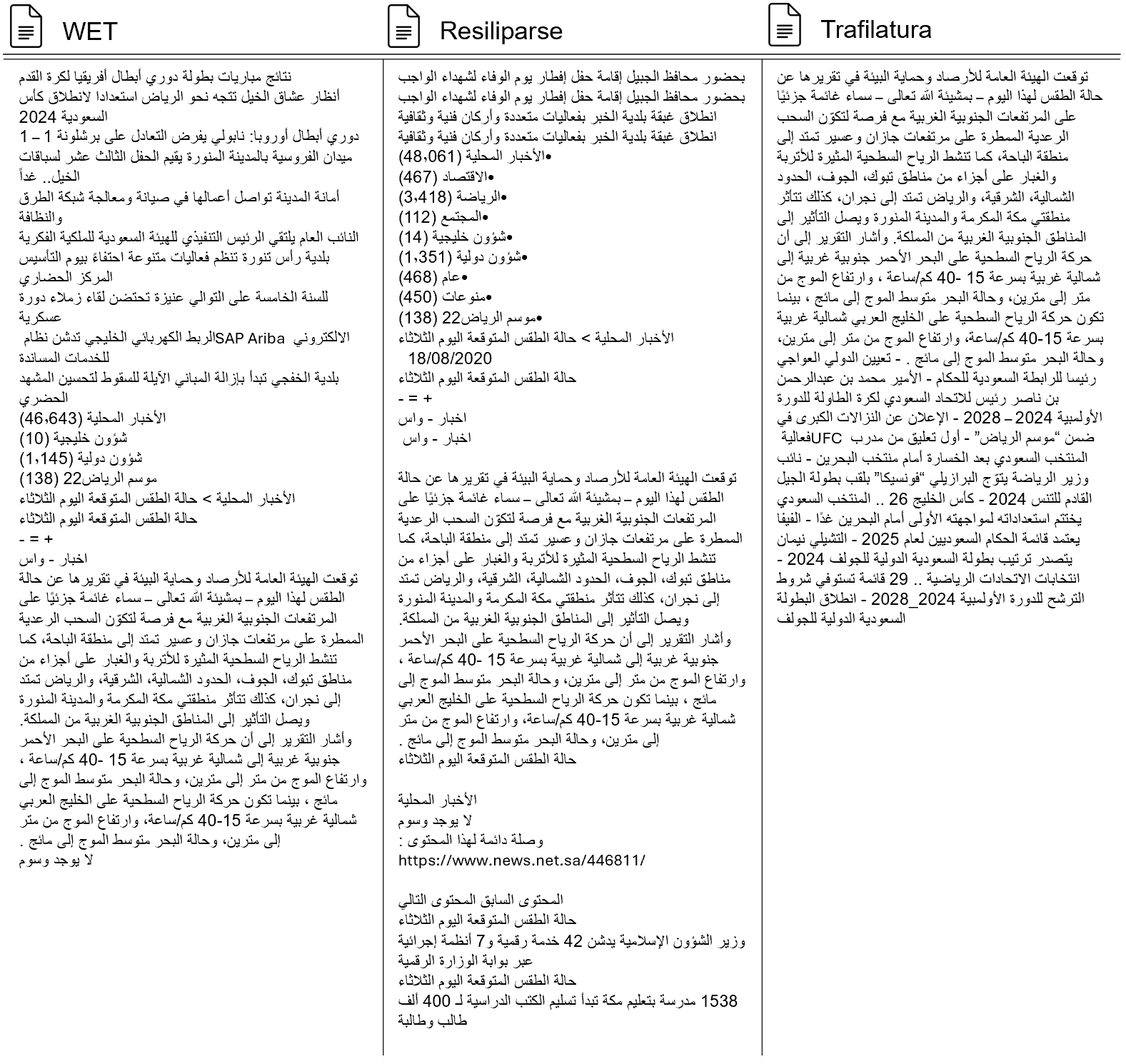}
    \vspace{-0.15cm}
    \caption{Comparison of extractors on a sampled web page. Left: WET plaintext includes text outside the main content. Middle: Resiliparse retains some boilerplate (navigation labels and counters). Right: Trafilatura yields the cleanest extraction. Note that Resiliparse text is truncated for space.}
    \label{fig:text extraction}
\end{figure}

 \section{Instructions for the Annotated Subset}
 As outlined in Section 2.2.4, this appendix presents the instructions provided to annotators for labeling the Arabic text samples. These guidelines ensured consistent evaluation of text clarity and readability across annotators.

    \makebox[\textwidth][c]{%
  \begin{minipage}{\textwidth}
    \centering
    \includegraphics[width=\textwidth,keepaspectratio]{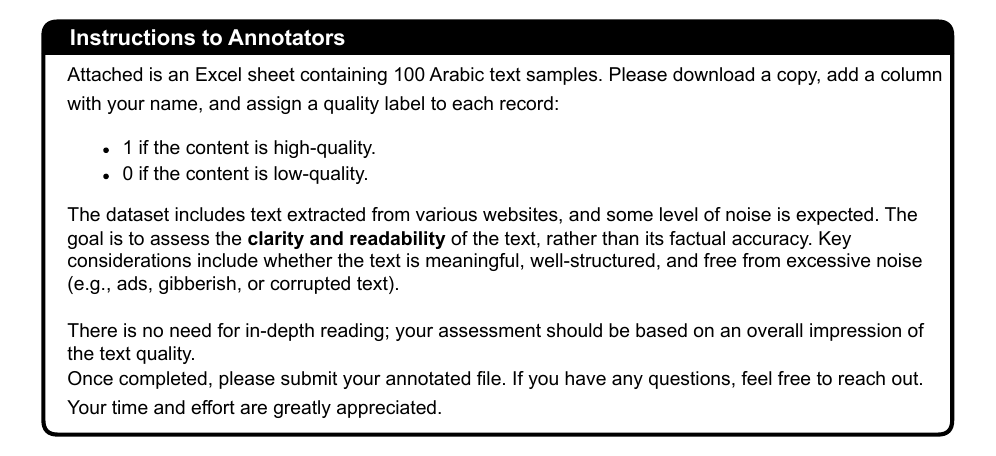}
    \captionof{figure}{Instructions provided to annotators for constructing the Annotated Subset. 
      Each annotator independently assigned binary quality labels (1 = high quality, 0 = low quality) to 100 text samples, 
      based on clarity, readability, and level of noise. 
      Final labels were aggregated via majority voting to create the ground truth for classifier evaluation.}
    \label{fig:annotator_instructions}
  \end{minipage}
}

\section{Translation Assessment Prompt}
Below is the prompt that guides the Gemini3.5 to evaluate Arabic translations of English questions across three stages: question translation, answer options translation, and overall classification. This structured process ensures accurate semantic alignment and helps identify translation issues.
 
\begin{figure}[h]
    \centering
    \includegraphics[width=1.01\linewidth]{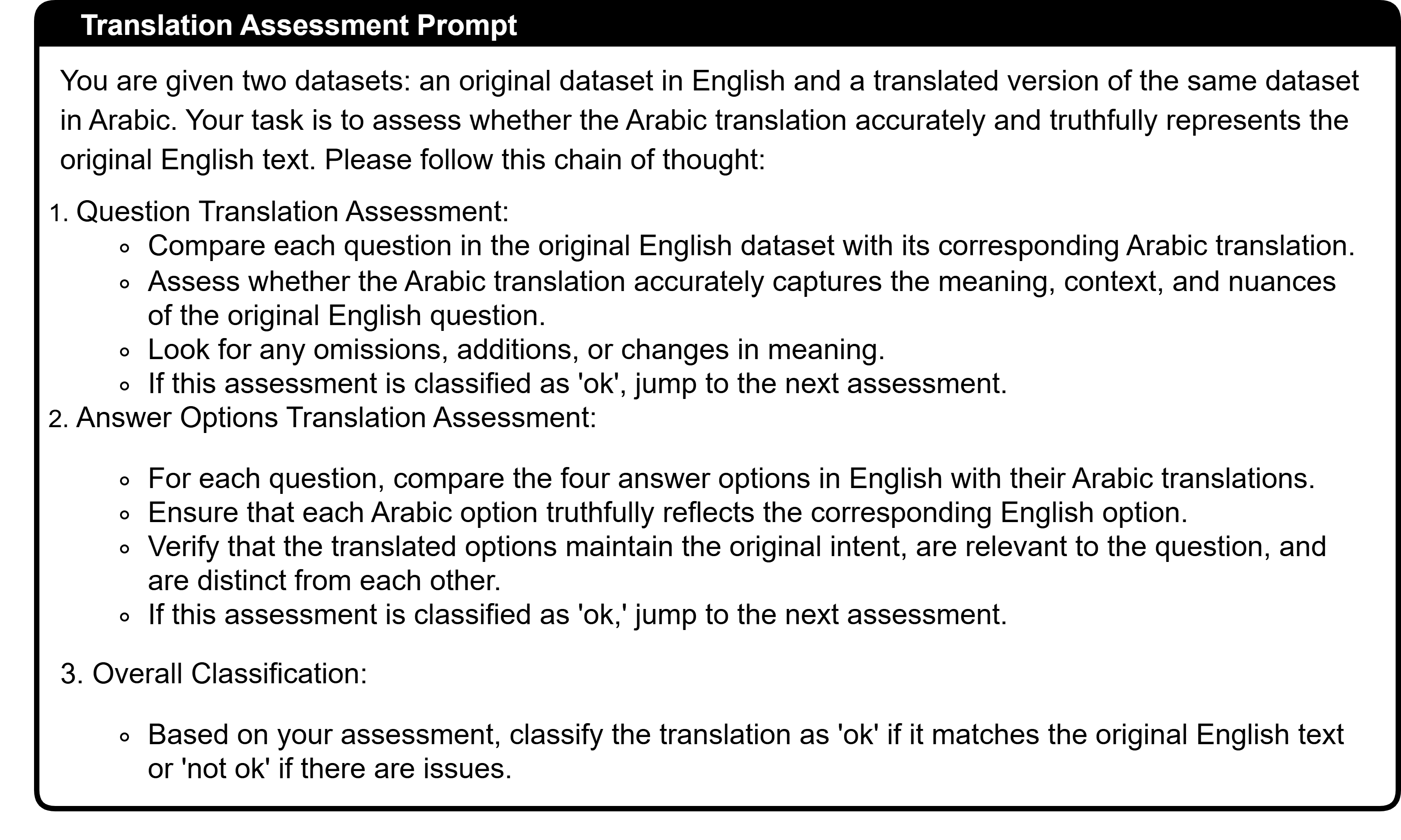}
    \label{fig:enter-label}
\end{figure}

\end{document}